%% file: preprint.tex
\PassOptionsToPackage{table}{xcolor}
\documentclass{article} 
\usepackage{iclr2025_conference,times}
\iclrfinalcopy
\input{math_commands}

\usepackage{hyperref}
\usepackage{url}
\usepackage{microtype}      
\usepackage{xcolor}         
\usepackage{subfigure}
\usepackage{url}
\usepackage{makecell}
\usepackage{multirow}
\usepackage{amsmath}
\usepackage{amssymb}
\usepackage{mathtools}
\usepackage{amsthm}
\usepackage{booktabs}
\usepackage{graphicx}
\usepackage{caption}
\usepackage{pifont} 
\usepackage{xspace}  
\usepackage{bbding}
\usepackage{MnSymbol}
\usepackage{dcolumn}
\usepackage{makecell}
\usepackage{threeparttable}
\usepackage{tikz}
\usepackage{algorithm}
\usepackage{algpseudocode}
\usepackage{adjustbox}
\usepackage{wrapfig}
\usepackage[table]{xcolor}
\usepackage{mdframed}
\usepackage{listings}
\usepackage{tcolorbox}
\tcbuselibrary{breakable}

\usepackage[utf8]{inputenc}

\theoremstyle{plain}
\newtheorem{theorem}{Theorem}[section]
\newtheorem{proposition}[theorem]{Proposition}
\newtheorem{lemma}[theorem]{Lemma}

\theoremstyle{definition}
\newtheorem{definition}[theorem]{Definition}

\theoremstyle{remark}

\def\methodname{Predictive-Decoding}

\title{Non-myopic Generation of Language Models for Reasoning and Planning}

\author{Chang Ma$^{\spadesuit}$ 
\quad Haiteng Zhao$^{\heartsuit}$ \quad 
\quad Junlei Zhang$^{\diamondsuit\Delta}$
\quad Junxian He\thanks{Equal advising.}$^{*\clubsuit}$
\quad Lingpeng Kong$^{*\spadesuit}$
 \\
$^\spadesuit$The University of Hong Kong \quad 
$^\heartsuit$Peking University 
\quad
$^\diamondsuit$Zhejiang University \quad \\$^\Delta$Westlake University \quad
$^\clubsuit$The Hong Kong University of Science and Technology 
\\\texttt{\{cma, lpk\}@cs.hku.hk, junxianh@cse.ust.hk}
}

\lstdefinelanguage{json}{
    showspaces=false,
    showtabs=false,
    breaklines=true,
    postbreak=\raisebox{0ex}[0ex][0ex]{\ensuremath{\color{gray}\hookrightarrow\space}},
    breakatwhitespace=true,
    basicstyle=\ttfamily\small,
    upquote=true,
    morestring=[b]",
    stringstyle=\color{black},
    literate=
     *{0}{{{\color{black}0}}}{1}
      {1}{{{\color{black}1}}}{1}
      {2}{{{\color{black}2}}}{1}
      {3}{{{\color{black}3}}}{1}
      {4}{{{\color{black}4}}}{1}
      {5}{{{\color{black}5}}}{1}
      {6}{{{\color{black}6}}}{1}
      {7}{{{\color{black}7}}}{1}
      {8}{{{\color{black}8}}}{1}
      {9}{{{\color{black}9}}}{1}
      {\{}{{{\color{black}{\{}}}}{1}
      {\}}{{{\color{black}{\}}}}}{1}
      {[}{{{\color{black}{[}}}}{1}
      {]}{{{\color{black}{]}}}}{1},
}

\newcommand{\gsm}{\textsc{GSM8K}\xspace}

\newcommand{\humaneval}{\textsc{HumanEval}\xspace}
\newcommand{\alfworld}{\textsc{AlfWorld}\xspace}

\newcommand\encircle[2][]{\tikz[overlay]\node[fill=blue!20,inner sep=2pt, anchor=text, rectangle, rounded corners=1.5mm,#1] {#2};\phantom{#2}}

\definecolor{multisample}{rgb}{0.94, 0.97, 1.0}
\definecolor{closed-loop}{RGB}{233, 196, 107}
\definecolor{scaling}{RGB}{130, 178,154}
\definecolor{global-optimal}{RGB}{246, 111, 105}
\definecolor{efficiency}{RGB}{033, 158,188}
\definecolor{strings}{RGB}{220, 20, 60}
\begin{document}

\maketitle

\begin{abstract}
\input{paragraphs/01_abstract_preprint}
\end{abstract}

\section{Introduction}
\input{paragraphs/02_introduction}

\section{Problem Formulation: an optimal control view\label{sec: preliminaries}}
\input{paragraphs/03_preliminaries}

\section{Beyond Myopic Generation for Better Planning\label{sec: predictive decoding}}
\input{paragraphs/04_method}

\section{Experiments}
\input{paragraphs/05_experiments}

\section{Related Work}
\input{paragraphs/07_related_work}

\input{paragraphs/final_parts}

\bibliography{ref}
\bibliographystyle{iclr2025_conference}

\appendix

\input{paragraphs/appendix}
\end{document}

%% file: math_commands.tex
\usepackage{amsmath,amsfonts,bm}

\def\eqref#1{equation~\ref{#1}}

\def\1{\bm{1}}

\def\vf{{\bm{f}}}

\DeclareMathAlphabet{\mathsfit}{\encodingdefault}{\sfdefault}{m}{sl}
\SetMathAlphabet{\mathsfit}{bold}{\encodingdefault}{\sfdefault}{bx}{n}

\DeclareMathOperator*{\argmax}{arg\,max}

%% file: paragraphs/01_abstract_preprint.tex
Large Language Models (LLMs) have demonstrated remarkable abilities in reasoning and planning by breaking down complex problems into sequential steps. Despite their success in various domains, such as mathematical problem-solving and coding, LLMs face challenges in ensuring reliable and optimal planning due to the inherent myopic nature of autoregressive decoding. This paper revisits LLM reasoning from an optimal control perspective, proposing a novel method, \methodname{}, that leverages Model Predictive Control to enhance planning accuracy. By reweighting LLM distributions based on foresight trajectories, \methodname{} aims to mitigate early errors and promote non-myopic planning. Our experiments show significant improvements across a wide range of tasks in math, coding, and agent-based scenarios. Furthermore, \methodname{} demonstrates computational efficiency, outperforming search baselines while utilizing inference compute more effectively. This study provides insights into optimizing LLM planning capabilities. Code is available at this \color{cyan}{\href{https://github.com/chang-github-00/LLM-Predictive-Decoding}{repo}}.

%% file: paragraphs/02_introduction.tex
Large Language Models (LLMs) are extensively pretrained on large corpus to predict the next tokens. Models like GPT-4~\citep{achiam2023gpt} have demonstrated a capacity for step-by-step reasoning and planning, breaking down complex problems into sequential steps that progressively lead to solutions~\citep{wei2022chain}. This \emph{sequential planning} capability has led to significant advancements in mathematical problem-solving \citep{cobbe2021gsm8k} and programming tasks \citep{chen2021evaluating}. It has also enabled essential applications such as tool use \citep{qin2023toolllm} and the development of LLM agents \citep{yao2022react}. Consider a scenario where an agent is tasked with cooking a meal: given an instruction like “\emph{Put salt on steak}” and an initial observation such as “\emph{There is a shelf in front of you},” the LLM assigns the highest probability to the most appropriate action. After executing this action, the LLM generates subsequent actions, each building upon the previous steps. This process creates a sequential chain of actions that advances toward task completion, guided by continuous interaction with environmental observations:
\begin{align*}
    &p (\text { ``\emph{Go to the shelf.}"} \mid \text{``\emph{There is a shelf in front of you.}"})\\  
    &p (\text { ``\emph{Pick up the saltbottle.}"} \mid \text { ``\emph{Go to the shelf. A saltbottle is on the shelf.}"}), 
\end{align*}
where ``\emph{A saltbottle is on the shelf}" is the new observation after going to the shelf. This iterative process continues until the specified goal state is achieved. However, this autoregressive decoding process could easily lead to irreversible errors, as LLMs tend to follow the most natural local follow-ups. For instance, given a different observation ``\emph{A spice bottle is on the shelf}'', the LLM is likely to generate actions ``\emph{Pick up the spice bottle}'' followed by ``\emph{Shake the bottle over the steak}''. Such a sequence of actions totally conflicts with the intended instruction ``\emph{Put salt on the steak}''. This contrasts with reinforcement learning paradigms, which use reward-based training to teach models to anticipate long-term action consequences and learn optimal decision policies~\citep{silver2014deterministic}. The short-sightedness of LLMs raises critical questions about their planning capabilities: (i) Can an LLM proactively avoid erroneous steps without necessitating their occurrence?
(ii) What is the degree of optimality achievable in LLM-based planning strategies?

\begin{figure}[!t]%
    \centering
    {\includegraphics[width=1.0\linewidth]{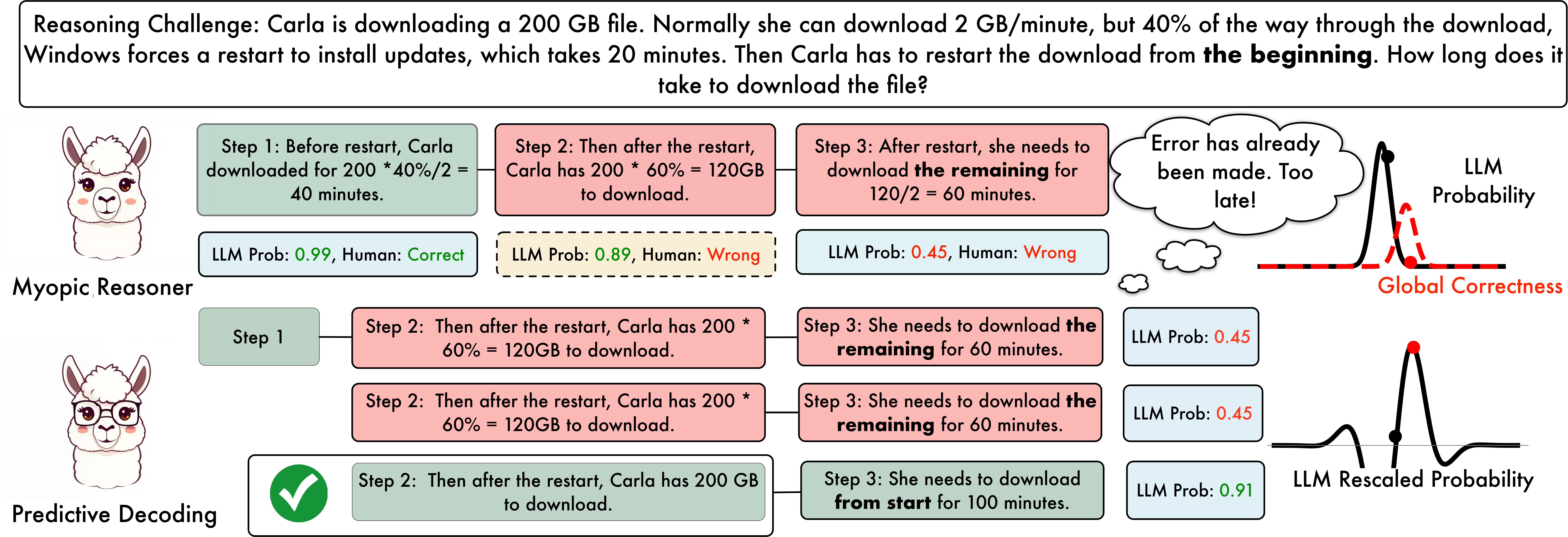} }%
    \caption{The illustrative overview of \methodname{} on one GSM8K example. LLM autoregressive planning often suffers from near sight. \methodname{} rescales LLM generation distribution based on evaluation of foresight, enabling non-myopic planning.}
    \label{fig:overall illustration}%
    \vspace{-0.5cm}
\end{figure}


To investigate the causes and extent of \underline{myopia} (short-sightedness) in LLM planning, we refer to optimal control theory \citep{qin1997overview}. Optimal planning is framed as a combinatorial optimization problem aimed at maximizing goal achievement through $T$ steps of actions $(a_1,a_2,\dots,a_T)$. Model Predictive Control (MPC; \citealp{camacho2007constrained}) outlines the optimality condition for  LLM-like planners -- sequential planning must be non-myopic, which requires the anticipation of future outcomes in addition to considering past actions, represented as $p(a_t \mid \text{\emph{History}}, \text{\emph{Future}})$. Conversely, focusing solely on $p(a_t \mid \text{\emph{History}})$ can lead to irreversible mistakes and potential planning failures. Recent studies~\citep{shih2023long,bachmann2024pitfalls} have observed myopia in language models pretrained with next-token prediction. Our research further reveals that even in advanced LLMs like \texttt{Llama-3}~\citep{dubey2024llama}, more than half of the reasoning processes on popular math tasks demonstrate a lack of global awareness, with myopia being particularly pronounced in incorrect instances (\S\ref{sec: finding 1}). Additionally, LLMs initially do not seem sensitive to mistakes; however, errors accumulate during multi-step planning and become increasingly noticeable over several steps (\S\ref{sec: finding 2}). 
These findings emphasize the importance of thinking with foresight.

To address this, we propose \methodname{}, a training-free approach to improve LLM planning with non-myopic generation. \methodname{} re-weights LLM distribution based on the closed-form solution in Model Predictive Control (Eq. \ref{eq: model predictive control}), such that sampling from the re-weighted solution is approximately solving planning with global-optimality. For each step $a_t$, \methodname{} samples multiple foresight trajectories $(a_t, a_{t+1},\dots, a_{t+T_0)}$ and rescales the original generation distribution based on the evaluations of foresight trajectories. Best action $a_t$ is sampled from this distribution. The process repeats until a solution is generated at timestep $T$. An overview is illustrated in Figure~\ref{fig:overall illustration}. Extensive experiments on math, coding, and agent tasks show notable improvements in planning and reasoning accuracy. For example, without using any additional supervision, we achieve a 7.2\% improvement on GSM8K~\citep{cobbe2021gsm8k} and a 25.3\% improvement on AlfWorld~\citep{ALFWorld20} accuracy over ReAct~\citep{yao2022react}. \methodname{}'s conditioning on future foresight promotes non-myopic planning and reduces early errors. It requires no special prompt design and preserves generation diversity. 

Furthermore, \methodname{} utilizes inference compute more effectively than other planning approaches. Previous methods have implemented various techniques to enhance planning global-awareness, such as weighted voting of diverse generations \citep{wang2022self} and searching algorithms \citep{yao2024tree, xie2024self, hao2023reasoning, zhou2023language}. While these methods can improve performance, they often require significantly higher computational resources, resulting in sub-optimal computation-performance trade-offs, as noted in recent studies \citep{snell2024scaling, wu2024empirical}. In contrast, \methodname{} adopts a ``no-regrets" approach that optimizes sequentially and simplifies the solution space, avoiding the exponential growth associated with iterative planning methods. This efficiency is reflected in our inference scaling law results with a fixed LLM and reward model, where \methodname{} achieves Pareto superiority over all state-of-the-art (SOTA) baselines (\S\ref{sec: inference scaling law}). Specifically, on the GSM8K benchmark, \methodname{} outperforms MCTS~\citep{hao2023reasoning}  by 2.4\% and Guided Decoding~\citep{xie2024self} by 1.4\% , while utilizing only about 50\% of the FLOPS, demonstrating its effectiveness and efficiency.

%% file: paragraphs/03_preliminaries.tex
One of the fundamental problems in decision-making is whether we can solve a long-term goal step by step~\citep{qin1997overview}. Traditionally, reinforcement learning approaches have relied on multiple iterations of training~\citep{silver2014deterministic} and searching~\citep{silver2017mastering} to achieve optimal results. However, recent advancements demonstrate that LLMs possess the capability for sequential planning~\citep{wei2022chain}. Sequential planning can achieve optimal outcomes if non-myopic, while also offering other benefits, as shown in Table~\ref{table: method comparison}. In this section, we first outline the context of the problem and reconsider LLM planning from the perspective of trajectory optimization.

\input{tables/comparison}

\paragraph{Planning as Trajectory Optimization} We consider Partially Observable Markov Decision Processes (POMDPs) defined by tuple $\langle g, \mathcal{S}, \mathcal{A}, \mathcal{T}, \mathcal{R}\rangle$, with goal $g$, state space $\mathcal{S}$, valid actions space $\mathcal{A}$, transition function $\mathcal{T}: \mathcal{S} \times \mathcal{A} \rightarrow \mathcal{S}$, and $\mathcal{R}: r(\mathbf{s},\mathbf{a}|g) \mapsto \mathbb{R}^+$ is the reward function that measures how well each step aligns with the goal (with the goal dependency omitted for simplicity).  \emph{Global-Optimal Planning} aims to find an action sequence that maximizes cumulative reward over a set number of steps. We model this as a Trajectory Optimization process:
find a sequence of actions $\mathbf{a}_{0: T}$ that maximizes an objective $\mathcal{J}$ (Return) factorized over per-timestep rewards $r\left(s_t, a_t\right)$ over a planning horizon with $T$ steps:
\begin{equation}
\begin{aligned}
     \underset{\mathbf{a}_{0: T}}\max&\mathcal{J}\left(\mathbf{s}_0, \mathbf{a}_{0: T}\right) = \sum_{t=0}^T r\left(\mathbf{s}_t, a_t\right), \\ 
    \text{ subject to } & \mathbf{s}_{t+1} = \mathcal{T}(\mathbf{s}_t, a_t), \ \ \forall t\in \left[0, T-1\right]
\end{aligned}
\label{eq: trajectory optimization original}
\end{equation}
Conversely, a \emph{Myopic Planner} maximizes the return on a shorter horizon. For instance, a planner that maximizes the immediate reward at current step, i.e $a_t = {\max}_{a_t}r(\mathbf{s}_t,a_t)$, is myopic, while \emph{non-myopic planning} works towards global-optimality.

Previous approaches to trajectory optimization follow a straightforward intuition: using large language models (LLMs) to generate diverse trajectories through sampling or search, and then selecting the best one \citep{wang2022self, yao2024tree, hao2023reasoning, xie2024self, shinn2024reflexion}. However, a significant drawback is the exponential growth of the solution space. Each action step offers $|\mathcal{A}|$ possibilities, leading to $|\mathcal{A}|^T$ potential solutions after $T$ steps. This exponential scaling complicates the optimization process, and despite attempts to prune trajectories and accelerate search \citep{zhuang2023toolchain, wang2024litesearch}, most iterative planning methods are either time-consuming or perform poorly with limited iterations \citep{chen2024tree}. Additionally, iterative methods are often unsuitable for closed-loop scenarios, where agents interact with their environment in real-time, as each decision alters the environment and is irreversible.

\paragraph{Sequential Planning with Model Predictive Control} Model Predictive Control~\citep{qin1997overview} introduces a different paradigm for planning: instead of optimizing an entire action sequence $\mathbf{a}_{0:T}^{\prime}$ at once, this method selects the best action at each timestep $a_0^{\prime}, \dots, a_t^{\prime}$, fixing each as it progresses and then optimizing the subsequent steps. This transforms Eq.~\ref{eq: trajectory optimization original} into a series of optimization problems, each with linear size solution space $|\mathcal{A}|$. The colors denote: \textcolor{red}{the current action to decide} $\color{red}a_t$, $\color{black} \text{\emph{history}}$, and $\color{blue}\text{\emph{future}}$.
\begin{equation}
\left\{\begin{array}{l}
a_0^{\prime}=\underset{\color{red}{a_0}}{\argmax} \underbrace{\left[\underset{\color{blue}{a_1,\dots,a_T}}{\max}\mathcal{J}\left(\mathbf{s}^0; \color{red}{a_0}, \color{blue}{a_1, \dots, a_T}\right)\right]}_{\text{Tail Subproblem}} \\
a_{1}^{\prime}=\underset{\color{red}{a_1}}{\argmax} \left[\underset{\color{blue}{a_2,\dots,a_T}}{\max}\mathcal{J}\left(\mathbf{s}^0, a_0^{\prime};\color{red}{a_1}, \color{blue}{\dots, a_T}\right)\right] \\
\cdots \\
a_{T}^{\prime}=\underset{\color{red}{a_T}}\argmax \left[\mathcal{J}\left( \mathbf{s}^0, \mathbf{a}_{0:T-1}^{\prime};\color{red}{a_T}\color{black}{}{}\right) \right]
\end{array}\right.
\label{eq: model predictive control}
\end{equation}

This method requires two steps: first, for each action $a_t$, compute the maximum future outcomes (tail subproblem), and then select the action that yields the best future outcomes. Model Predictive Control can also use sampling for inference, but it improves sampling efficiency in two folds: the solution space for each equation is linear and the size of the tail subproblem is reduced after each step. We further provide sample efficiency theoretical analysis and simulations in Appendix \ref{sec: sample efficiency}. 


\paragraph{LLM Autoregressive Planning} Solving Eq. \ref{eq: model predictive control} is challenging -- We need to identify an action that achieves the best outcome after $T$ steps. LLM provides a straightforward solution to this problem: it leverages its extensive knowledge to anticipate the next best action, $a_t^{\prime}$, for achieving the goal. This method can be formalized as: 
\begin{equation}
\left\{\begin{array}{l}
a_0^{\prime} =\underset{a_0}{\arg\max} P^{\text{LLM}}\left(a_0\mid \mathbf{s}_0\right) \\
a_{1}^{\prime}= \underset{a_1}{\arg\max} P^{\text{LLM}}\left(a_1\mid \mathbf{s}_0, a_0^{\prime}\right) \\
\cdots \\
a_{T}^{\prime} = \underset{a_T}{\arg\max} P^{\text{LLM}}\left(a_T\mid \mathbf{s}_0, \mathbf{a}_{0:T-1}^{\prime}\right) 
\end{array}\right.
\label{eq: LLM}
\end{equation}
This approach is widely adopted in LLM reasoning and planning, encompassing Chain-of-Thought (CoT; \citealt{wei2022chain}), ReAct \citep{yao2022react}, and Voyager \citep{wang2023voyager}. Research has demonstrated that with carefully designed prompts and in-context examples, LLM autoregressive planning can achieve competitive performance in various tasks \citep{cobbe2021gsm8k, chen2021evaluating, ALFWorld20}. We further investigate whether LLM autoregressive planning is global-optimal.

\section{Diagnosing the Deficiency of LLM Planners}
Comparing Eq.\ref{eq: LLM} with Eq.\ref{eq: model predictive control}, we observe that LLM-based action selection at time step $t$ differs from the optimal control approach. Instead of maximizing future outcomes in the tail subproblem, LLMs select actions based on the immediate conditional probability $\argmax_{a_t}P^{\text{LLM}}(a_t\mid \mathbf{s}^0, a_{0:t-1}^{\prime})$.\footnote{Each token within $a_t$ is generated based on the highest current probability; we express this at the action level for brevity.} Consequently, the optimality of LLM planning hinges on two hypotheses: (1) \textit{Non-myopic}: LLMs inherently plan ahead, with current step probabilities accounting for the success of future steps. (2) \textit{Evaluation capability}: The probability distribution in LLMs can effectively substitute for $\mathcal{J}$, distinguishing successful trajectories from failed ones. In the following sections, we will examine and investigate the validity of these two hypotheses.

\subsection{Finding 1: LLM Autoregressive Planning is Myopic\label{sec: finding 1}} 
We first examine whether LLM inherently plan ahead. Next token prediction is greedy; however previous evidence suggests that extensive pretraining could allow LLMs to implicitly plan ahead for future tokens~\citep{wu2024language}. Here we analyze the myopic gap for LLM planning.

\begin{definition}\textit{(Myopic Gap for LLM Planning)} Given a language model with distribution $P(a_0,a_1,\dots, a_T)$, let $\mathcal{P}$ be the support set of the distribution. ${\mathbf{a}}_{0:T}'$ are generated autoregressively following Eq.~\ref{eq: LLM}. Then the myopic gap for planning is: 
\begin{equation}
    p^{\ast} = \underset{\mathbf{a}_{0:T}\in \mathcal{P}}{\max}P(a_0,a_1,\dots,a_T) - P({a'_0},{a'_1},\dots,{a'_T}) 
\end{equation}
\end{definition}

$p^*$ represents the difference between $P^{\text{LLM}}(a_t \mid \text{\emph{history}},\text{\emph{future}})$ and $P^{\text{LLM}}(a_t \mid \text{\emph{history}})$. A positive gap ($p^* > 0$) indicates that the LLM's action selection is myopic for at least one intermediate step, while $p^*=0$ proves consistent global-awareness during planning. Here we include a small margin ($p^*>0.01 \Rightarrow$ myopic) to discriminate myopic and non-myopic cases.

\begin{figure}[t!]  
  \centering   
  \begin{minipage}{0.25\textwidth}   \includegraphics[width=\textwidth]{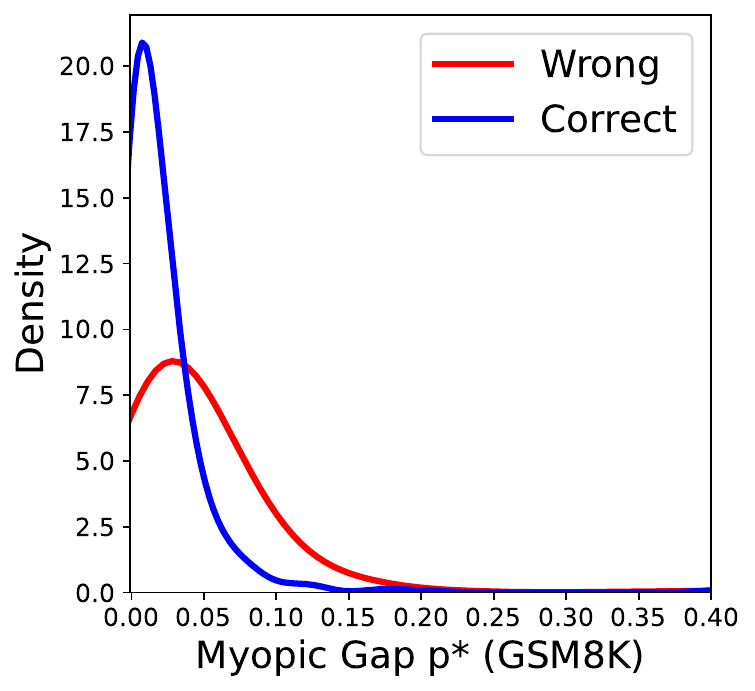}
  \end{minipage}
  \begin{minipage}{0.25\textwidth} 
  \includegraphics[width=\textwidth]{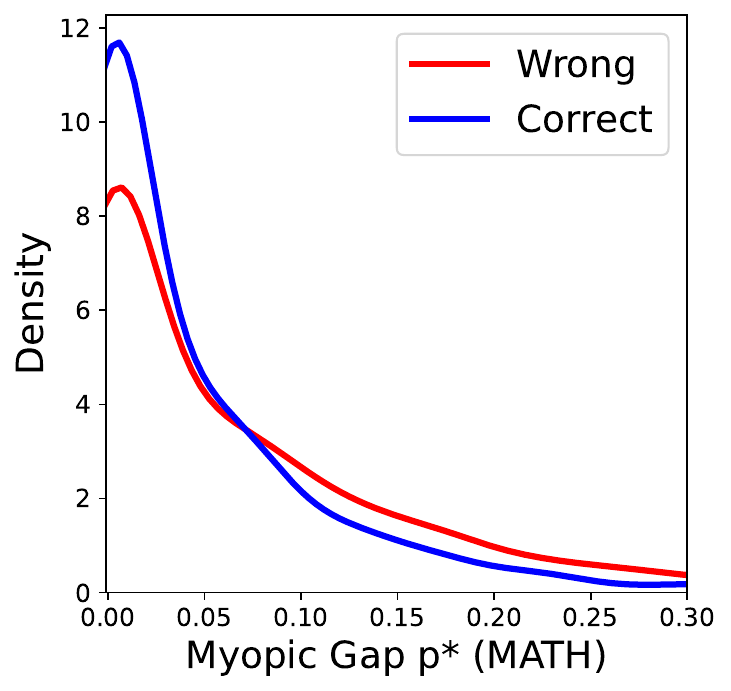}
  \end{minipage}
   \hspace{0.035\textwidth}
  \begin{minipage}{0.33\textwidth}  
  \centering
  \vspace{-0.35cm}
  \resizebox{\linewidth}{!}{
      \begin{tabular}{lccc}
\toprule
            & Wrong  & Correct & Overall \\
                    \midrule           
\textit{GSM8K} & &  \\
Myopic      & 290   & 539  & 829  \\
Non-myopic  & 89   & 401  & 490 \\
Myopic Rate & \textbf{76.5}\% & 57.3\% & 62.8\% \\
\midrule
\textit{MATH} & &  \\
Myopic      & 2172   & 785   &  2947 \\
Non-myopic  & 1341   & 702   &   2043\\
Myopic Rate & \textbf{61.8}\% & 52.7\%  & 59.1\% \\
\bottomrule
\end{tabular}
        }
  \end{minipage} 
  \caption{Myopic Gap distribution for correct and wrong samples (drawn with kde-plot). Myopic examples are defined as $p^*>0.01$. Wrong samples show a higher myopic rate on both tasks. }
  \label{fig: confusion matrix}
  \vspace{-0.5cm}
\end{figure}

We empirically calculate the myopic gap of \texttt{Llama3-8B} on two challenging reasoning tasks, GSM8K~\citep{cobbe2021gsm8k} and MATH~\citep{hendrycks2021measuring}. We ensemble the generation of token-level Beam Search~\citep{freitag2017beam} and Predictive Decoding (detailed in \S\ref{sec: predictive decoding}) as the non-myopic target. Results in Figure \ref{fig: confusion matrix} show that around 60\% of all examples show are myopic. Additionally, correct samples are often less myopic, while 76.5\% of GSM8K errors and 61.8\% of MATH errors occur when LLM planning is myopic. Correct samples tend to have smaller gaps, this motivates us that optimizing generation towards non-myopic may lead to improved results. 


\subsection{Finding 2: LLM struggles to identify mistakes in planning early\label{sec: finding 2}} 


In this section, we explore another bottleneck in LLM sequential planning:  Can LLMs evaluate intermediate steps and identify mistakes early on?

\begin{wrapfigure}{r}{0.4\textwidth}  
    \centering  
    \vspace{-5mm}
    \includegraphics[width=\linewidth]{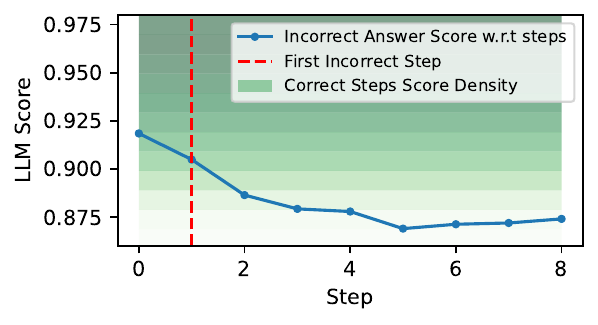}
    \vspace{-5mm}
    \caption{In GSM8K, the first incorrect step's average score is among correct steps, but not after a few steps.  }  
    \label{fig: logp compare}  
    \vspace{-5mm}
\end{wrapfigure} 

We perform an LLM score calibration analysis by comparing LLM evaluations of intermediate steps with ground truth human annotations. We collect samples of trajectories $a_{0:t}, t\in (0,T]$ and use the LLM to evaluate these trajectories $P^{\text{LLM}}(a_0,a_1,\dots, a_t)$. \texttt{GPT-3.5-Turbo} is used to evaluate trajectories on the agent task AlfWorld, while \texttt{Llama3-8B} evaluates steps of the GSM8K trajectory. The LLM score for Llama3 is calculated using LLM probability, and the score for proprietary GPT-3.5 is obtained through prompt-based self-evaluation~\citep{xie2024self}. Human annotators label each step in the AlfWorld trajectory with scores from ${0, 0.25, 0.5, 0.75, 1}$, where $0$ indicates that the task is unlikely to be completed and $1$ indicates a high likelihood of successful completion. Each step in the GSM8K trajectory is labeled as either Correct or Incorrect following ~\citet{lightman2023let}.


As shown in Table \ref{tab: reward calibration}, \texttt{GPT-3.5-Turbo} could barely estimate intermediate progress simply with the current trajectory $\mathbf{a}_{0:t}$, with only $\rho=0.133$ correlation to human ground truth and Calibration Error 0.165. However, after given future trajectory (foresight of 3 steps), LLM evaluation improves.
This is in line with previous observations~\citep{uesato2022solving, lightman2023let} that LLMs are more natural at evaluating complete trajectories, rather than intermediate actions.

The evaluation of intermediate steps for \texttt{Llama3-8B} on GSM8K shows a calibration error of 0.332 when compared to human ground truth. We observe that incorporating foresight into the evaluation could help avoid early mistakes. As shown in Figure \ref{fig: logp compare}, the score for the first incorrect step falls within the high-density region of correct step scores but becomes easily distinguishable from correct steps after a few iterations. Consequently, the GSM8K calibration score decreases to 0.323 when foresight of three steps is applied (detailed in Table ~\ref{tab: reward calibration}). However, the correlation does not improve, as the scores of positive steps slightly drops, leading to some overlap with ambiguous incorrect steps.



\input{tables/calibration_reward}

%% file: tables/comparison.tex
\begin{table*}[!t]
    \centering
    \caption{Non-iterative planning is efficient and suitable for closed-loop settings. Iterative planning, though better for global optimal decisions, is slow and impractical for realistic agent tasks. \methodname{} combines the best of both. Advantages include: \encircle[fill=global-optimal, text=white]{G}\ lobal-Optimality of solution,  \encircle[fill=scaling, text=white]{S}\ caling to long-range planning,  \encircle[fill=efficiency, text=white]{E}\ fficiency, and \encircle[fill=closed-loop, text=white]{C}\ losed-Loop Interaction.}
    \resizebox{\linewidth}{!}{
    \begin{tabular}{lcccl}
        \toprule
        \textbf{Decision Making} &\textbf{Paradigm Illustration} & \makecell{\textbf{Representative Methods}}& \textbf{Type}&\textbf{Advantage}
        \\
        \midrule
        \multirow{5}{*}{\textbf{Iterative Planning}} & \multirow{5}{*}{\includegraphics[height=50pt]{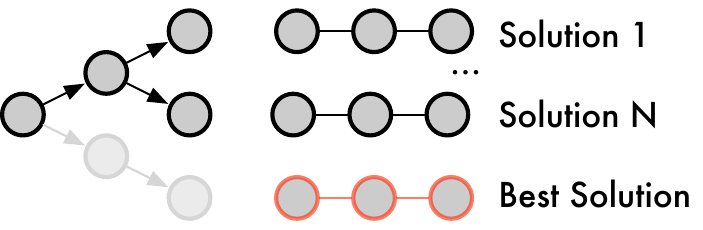}}&Self-Consistency~\citep{wang2022self} & Sample & \multirow{5}{*}{\encircle[fill=global-optimal, text=white]{G}\hspace{1 mm} }\\
        && {Tree-of-Thought~\citep{yao2024tree}} &\textcolor{gray}{Search} &   \\
        && {Guided-Decoding~\citep{xie2024self}}&\textcolor{gray}{Search} &   
        \\
        && {Monte Carlo Tree Search~\citep{hao2023reasoning}}&\textcolor{gray}{Search} &  \\
        && {Reflexion~\citep{shinn2024reflexion}} & \textcolor{brown}{Prompt} &  
        \\    
        \midrule 
        \multirow{3.5}{*}{\textbf{Sequential Planning}}& \multirow{3.5}{*}{\includegraphics[height=27pt]{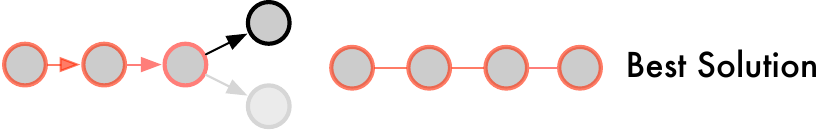}} & Chain-of-Thought~\citep{wei2022chain} & Sample & \multirow{2}{*}{\encircle[fill=scaling, text=white]{S} \hspace{0.2 mm} \encircle[fill=efficiency, text=white]{E}\hspace{1.2 mm}
        \encircle[fill=closed-loop, text=white]{C}}
        \\
        && {ReAct~\citep{yao2022react}} & Sample &  \\
        \cmidrule{3-5}
        && \methodname{} (Ours) & Sample & \encircle[fill=global-optimal, text=white]{G}\hspace{1 mm} \encircle[fill=scaling, text=white]{S} \hspace{0.2 mm} \encircle[fill=efficiency, text=white]{E}\hspace{1.2 mm} 
        \encircle[fill=closed-loop, text=white]{C}\\
         \bottomrule
    \end{tabular}
    }
    \label{table: method comparison}
    \vspace{-0.5cm}
\end{table*}

%% file: tables/calibration_reward.tex

\begin{table}[ht!]
\centering
\caption{LLM Score Calibration vs. Human Ground Truth.$^\ddagger$ECE for AlfWorld uses a 0.5 threshold for binary classification.}
\resizebox{0.9\linewidth}{!}{
\begin{tabular}{cccccc}
\toprule
\multirow{2}{*}{Task} & \multirow{2}{*}{Model}  & Calibration Error ECE  & Calibration Error ECE & Correlation $\rho$ & Correlation $\rho$   \\ &  & (wo. Foresight) & (w. Foresight) $\color{scaling}\downarrow$  & (wo. Foresight) & (w. Foresight) $\color{scaling}\uparrow$ \\
\midrule
AlfWorld
 &    {\textsc{GPT-3.5-Turbo}}    & 0.165$^\ddagger$ & 0.108$^\ddagger$ $\color{scaling}\downarrow$ & 0.133 & 0.291 $\color{scaling}\uparrow$
\\
GSM8K &   \textsc{Llama-3} &    0.332 & 0.323 $\color{scaling}\downarrow$ & 0.417 & 0.376  $\color{global-optimal}\downarrow$     \\ 
\bottomrule
\end{tabular}
}
\label{tab: reward calibration}
\end{table}

%% file: paragraphs/04_method.tex

Our findings presented above highlight critical issues with current LLM decoding methods, particularly their tendency toward myopic planning. Additionally, the findings suggest that incorporating global planning strategies can significantly enhance performance. By incorporating future generations, we can calibrate erroneous actions, which directly motivates our proposed method. In this section, we introduce \methodname{}, which follows MPC principles to reduce myopic planning in LLMs. The pipeline is introduced in following sections and the pseudocode is presented in Algorithm \ref{alg:mpc} in Appendix \ref{sec: pseudocode}.

\subsection{\methodname\label{sec: method main}}
Our decision-making strategy draws inspiration from model predictive control (MPC) ~\citep{qin1997overview}. It solves the $T$ sub-optimization problems (Eq. \ref{eq: model predictive control}) and addresses myopia with foresight. Here in order to ensure non-myopic planning, our main objective is to generate $a_t^{\prime}$ according to:
\begin{equation}\label{eq: solution}
a_t^{\prime}=\underset{a_t}{\arg \max}\ \mathbb{E}_{\mathbf{a}_{>t}, \mathbf{s}_{>t}} P^{\text{LLM}}\left(a_t, \mathbf{a}_{>t}, \mathbf{s}_{>t} \mid \mathbf{a}_{<t}^{\prime}, \mathbf{s}_{<t}^{\prime}\right), \forall a_t\in \mathcal{P},
\end{equation}

where $\mathcal{P}$ is the support set of distribution $P^{\text{LLM}}(a_t \mid \mathbf{a}'_{<t}, \mathbf{s}'_{<t})$,  and $\mathbf{a}_{>t}\sim  P^{\text{LLM}}(\cdot\mid \mathbf{a}'_{<t}, \mathbf{s}'_{<t}, a_t)$. Note that based on results from \S\ref{sec: finding 2}, LLMs could accurately evaluate future steps after incorporating a few steps of foresight. Therefore here we evaluate the future constrained to $T_0$ steps of foresight, i.e. $\mathbf{a}_{>t}:=\mathbf{a}_{t+1: t+T_0}$. 

We use constrained decoding and sampling to apply soft constraints on outputs, promoting diverse generation and preventing overfitting to our partial ($T_0$ steps) foresight. Our method follows the sampling-importance-resampling (SIR) technique from energy-based models (EBM)~\citep{smith1992bayesian,ji2023language} to achieve the optimization goal.

\begin{proposition}\label{thm: main}
    The distribution that solves the optimization problem in Eq.\ref{eq: solution} is in the form of: 
    \begin{equation}\label{eq: main}
        p_{\tau}(a_t)\propto P^{\text{LLM}}\left(a_t\mid \mathbf{a}_{<t}^{\prime}, \mathbf{s}_{<t}^{\prime}\right)\exp\left[ \mathbb{E}_{\mathbf{a}_{>t},\mathbf{s}_{>t}}P^{\text{LLM}}\left(a_t, \mathbf{a}_{>t}, \mathbf{s}_{>t} \mid \mathbf{a}_{<t}^{\prime}, \mathbf{s}_{<t}^{\prime}\right)/\tau\right].
    \end{equation}
\end{proposition}
The detailed proof is in Appendix~\ref{sec: proof 4.1}. Specifically we first use the LLM to sample a set with size K of foresight rollouts $\mathbf{a}_{t:t+T_0}$ given prefix $\mathbf{a}_{<t}^{\prime}$ in parallel, as well as obtaining the exponential value of generation probability $w_k$ of each rollout. Then we obtain the categorical distribution $\text { Categorical }\left(\frac{w_1}{\sum_{k=1}^K w_k}, \cdots, \frac{w_K}{\sum_{k=1}^K w_k}\right)$. The next action step is determined as the first step in the sampled rollout from this distribution. In the limit 
$K\rightarrow\infty$ and temperature $\tau \rightarrow 0$, the method recovers the exact maximum value of the distribution. In practice, we can choose 
$K$according to computational budget and set $\tau$ 
 according to whether we want the model to fit more towards the distribution or generate more diversely. The computation overhead of this method mainly stems from the tokens decoded additionally for foresight. The parallel sampling of multiple rollouts is generally fast for LLM inference infrastructures~\citep{kwon2023vllm}.

In previous studies, generating text with a lookahead approach has been applied to controllable language model generation \citep{deng2020residual, lu2021neurologic, qin2022cold}. These methods typically involve generating extra tokens to verify if constraints are met, accepting previous generations more likely to satisfy future constraints. Our approach builds on these work, focusing on future generations at the action level rather than the token level, better aligning with large language models' needs for reasoning and planning. While \citep{deng2020residual, qin2022cold} also use EBM sampling, our method is tailored for pretrained models without additional training. Unlike controllable generation, which often requires task-specific heuristics, we demonstrate that lookahead decoding alone can enhance performance in complex planning.

\methodname{} mainly uses LLM self-evaluation of foresight to improve planning, but this method could also be integrated with an external heuristics $J$. Eq.~\ref{eq: main} can be rewritten as $
    p(a_t)\propto P^{\text{LLM}}\left(a_t\mid \mathbf{a}_{<t}, \mathbf{s}_{<t}\right)\exp\left[ \mathbb{E}_{\mathbf{a}_{>t}}\mathcal{J}\left(\mathbf{s}_0;\mathbf{a}_{0:T}\right )\right]$, which uses the heuristic function $J$ rather than LLM probability to evaluate the foresight. This setting enables LLM to utilize discrete heuristics more effectively and makes it easier to compare with other search baselines which often use external reward models. More details are elaborated in \S\ref{sec: reward optimization experiment}.


\subsection{Recycling Trajectory Rollouts \includegraphics[height=10pt]{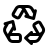}\label{sec: trajectory recycle}}

In previous work, foresight in generation has been applied for controllable tasks \citep{deng2020residual,lu2021neurologic}, typically using token-level constraints with a short lookahead of only a few tokens. In contrast, generating future actions involves creating $K$ samples with longer foresights, which could lead to increased computation with complexity $O(KT_0)$. Moreover, insufficient sampling number $K$ can inaccurately model the distribution $p_{\tau}(a_t)$, while larger $K$ is inefficient. Inspired by work on accelerating LLM inference~\citep{fu2024break}, we design a memory pool to recycle sampled trajectories that could reduce the number of sampling $K$. 

We observe that different sampled trajectories at different time-steps often overlap. For example, when an agent needs to heat a tomato, the trajectory starting from ``\textit{go to the microwave $\rightarrow$ put the tomato in a microwave $\rightarrow$ heat the tomato with microwave} " overlaps with ``open the fridge $\rightarrow$ take the tomato from fridge $\rightarrow$ \textit{go to the microwave $\rightarrow$ put the tomato in a microwave $\rightarrow$ heat the tomato with microwave}" since the tomato needs heating after being taken from the fridge. Thus, previously sampled trajectories and their probabilities $(\mathbf{a}_{t:t+T_0}, p_t)$ can be reused for evaluating timestep $t'\in (t, t+T_0)$, provided their trajectory matches $\mathbf{a}_{t:t'}=\mathbf{a}_{t:t'}^{\prime}$. Similarly, this could be applied for mathematical reasoning as well as coding, as separate reasoning paths often share steps. 

In implementation, we aggregate all sampled foresights in a trajectory pool. For deciding each new action, we retrieve trajectories with the correct history from the pool and perform the second ``resampling" stage based on these valid trajectories. This method recycles trajectories used in estimating foresight, enhancing sampling accuracy.

%% file: paragraphs/05_experiments.tex
In our experiments, we evaluate \methodname{} in two settings: (i) \textbf{without additional supervision}, using only LLM self-evaluation (\S\ref{sec: predictive decoding experiment}); (ii) \textbf{with an additional reward} (heuristics or reward model) to analyze \methodname{}'s sample efficiency in trajectory optimization (\S\ref{sec: reward optimization experiment}).

\subsection{Experiment Settings}

\textbf{Benchmarks} Our evaluation covers three domains: \emph{math}, \emph{coding}, and \emph{agents}. Math tasks GSM8K~\citep{cobbe2021gsm8k} and MATH~\citep{hendrycks2021measuring} are essential reasoning benchmarks, while coding tasks HumanEval~\citep{chen2021evaluating} and MBPP~\citep{austin2021program} are also closely related to reasoning ability. We also evaluate on two agent tasks AlfWorld~\citep{ALFWorld20} and PDDL (from Agentboard, \citealp{ma2024agentboard}) to understand planning ability in closed-loop interactions. 

\textbf{Evaluation Settings}
We use \texttt{Llama3}~\citep{dubey2024llama} and \texttt{Mistral-v0.3}~\citep{jiang2023mistral} for math tasks, and \texttt{Deepseek-Coder}~\citep{guo2024deepseek} for coding tasks. For agent planning which requires stronger models, we include the proprietary \texttt{GPT-3.5}~\citep{achiam2023gpt} alongside \texttt{Llama3.1-70B}. We use vLLM inference infrastructure for efficiency~\citep{kwon2023vllm}. To ensure fair comparisons, we use standardized prompts from ~\citet{guo2024deepseek}, ~\citet{gao2023pal}, ~\citet{cobbe2021gsm8k}, and \citet{ma2024agentboard}.  Hyperparameter and prompt details are stated in Appendix~\ref{sec: hyperparameters}


\textbf{FLOPS calculation} Many of our experiments report computational efficiency to better illustrate performance-efficiency tradeoff. We using FLOPS as the metric, ~\citet{kaplan2020scaling}, i.e. $\text{FLOPS}\approx 6nP$, where $P$ is the number of parameters in the LLM, and $n$ the number of generated tokens. We record the average number of generated tokens per example for this calculation.

\textbf{Baselines} We incoporate iterative planning baselines -- Self-Consistency~\citep{wang2022self}, Guided-Decoding~\citep{xie2024self}, MCTS~\citep{hao2023reasoning} and Self-Infilling~\citep{zheng2023self} -- as well as sequential planning baselines -- 
 ReAct~\citep{yao2022react}, Beam Search, PAL~\citep{gao2023pal}, COT~\citep{wei2022chain}) -- in our evaluation. We excluded search agents such as LATS \citep{zhou2023language} from our comparison, as they primarily address non-environment altering agents (e.g., QA and Retrieval agents).

\input{tables/reasoning}

\input{tables/code}

\subsection{Predictive Decoding Main Results\label{sec: predictive decoding experiment}}

\textbf{Results on Math Tasks} Following \citep{xie2024self}, we use the Program Aided Language Model (PAL) format for each step, and execute the generated code in a Python environment to obtain answers. As shown in Table \ref{tab: math main results}, \methodname{} significantly improves problem-solving accuracy: improve by 7.2\% on GSM8K and 4.3\% on MATH with \texttt{Llama3}, and 13.3\% and 6.8\% with \texttt{Mistral-0.3}. Beam Search achieves similar gains, which reduces myopia but requires over 3× the computation. Our method matches or exceeds Beam Search performance and can be further enhanced with Self-Consistency, outperforming both autoregressive and beam search methods in that setting.
Guided-Decoding performance well using 
\texttt{Codex}, but its performance drops using \texttt{Llama3} due to its overconfident confidence scores. Its performance improves when guided by a reward model (in \S~\ref{sec: reward optimization experiment}).

\textbf{Results on Coding Tasks} 
Table \ref{tab: coding main results} presents results for two code generation tasks evaluated using Pass@1 and Pass@10 settings. Pass@10 selects the best from 10 generations, highlighting both accuracy and diversity. \methodname{} outperforms all baselines with \texttt{Llama3}, achieving a strong quality-diversity balance, as detailed in the subsequent temperature analysis. Interestingly, beam search shows less stability in the Pass@10 setting for coding tasks, suggesting that search-based methods may result in lower diversity in generation.

\input{tables/agent}
\input{tables/ablation}

\textbf{Results on Agent Tasks}
Table \ref{tab: agent main results} presents the performance of two agent tasks: \methodname{} shows improvements of up to 24.6\% on two agent tasks and outperforms consistently over Act and ReAct. Notably, \methodname{} achieved more significant enhancements in success rates than in progress rates. While progress rates often reflect an agent's exploration skills, \methodname{} excels in strategic planning and error reduction, leading to higher success rates. Case studies compared to baselines are available in Appendix \ref{sec: case study}.

\textbf{\textit{Impact} of Foresight Length $T_0$} We analyze the impact of foresight length in Table \ref{tab: ablation study}. Performances on both GSM8K and HumanEval both improve with longer foresight, supporting our premise that non-myopic generation enhances planning. Notably, \texttt{Llama3}'s myopia on GSM8K is mitigated with just 4 foresight steps, indicating that LLMs already possess partial capability to plan ahead.

\begin{figure}[!t]

    \begin{minipage}{0.45\textwidth} 
    \centering
 \includegraphics[width=\linewidth]{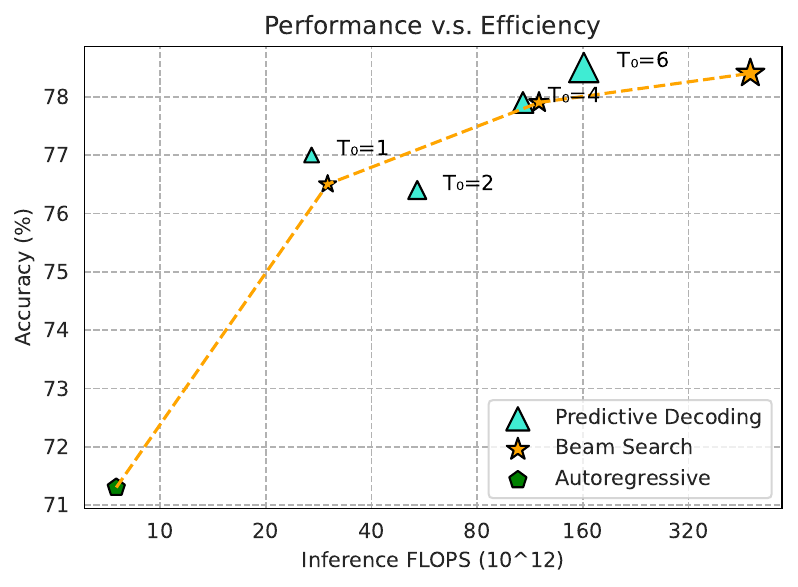}
 \vspace{-0.5cm}
\caption{Performance v.s. Efficiency on GSM8K. Predictive Decoding is Pareto superior to Beam Search with longer foresight.}
    \label{fig:gsm8k flops analysis}
    \end{minipage}
  \hfill
    \begin{minipage}{0.48\textwidth}
    \centering
\includegraphics[width=\linewidth]{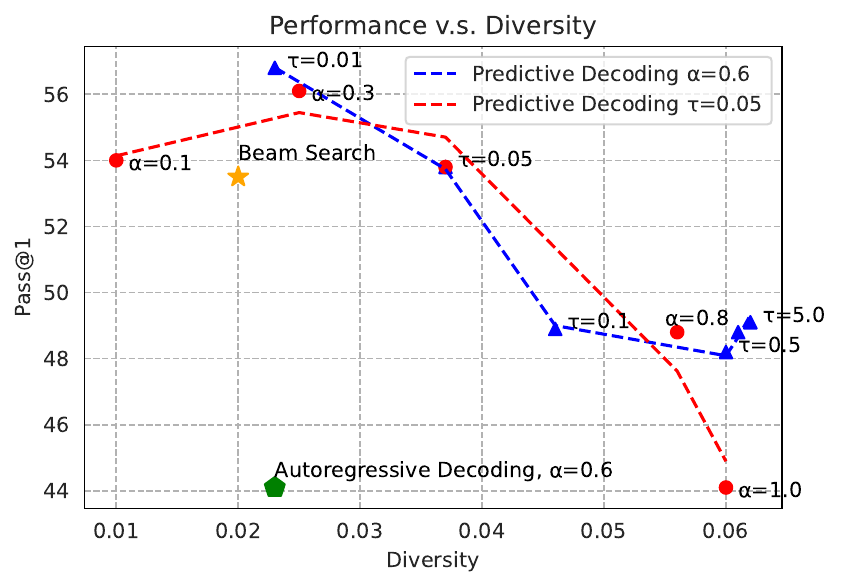}
\vspace{-0.5cm}
    \caption{Illustrating Performance and Diversity tradeoff on HumanEval by controlling parameters $\tau$ and $\alpha$. Diversity uses 1- ROUGE score.}
    \label{fig:humaneval temperature analysis}
    \end{minipage}
    \vspace{-0.3cm}
\end{figure}

On HumanEval, \methodname{} scale significantly with increased foresight length. This aligns with findings from studies on inference time scaling~\citep{snell2024scaling}, which we discuss further in later sections. Our method is more efficient in mitigating myopia compared to Beam Search, as shown in Figure \ref{fig:gsm8k flops analysis}. We achieve this with lower FLOPS, since beam search does not directly optimize for myopia. We further discuss inference compute scaling in \S\ref{sec: inference scaling law}.

\textbf{\textit{Impact} of Sampling number $K$ and Trajectory Recycle} We must sample from the original distribution as described in \S\ref{sec: method main}, but more sampling may reduce efficiency.
 As shown in Table \ref{tab: ablation study}, increasing $K$ leads to better performances as it could give a more accurate account of the original generation distribution. Also, trajectory recycle significantly improve the sampling efficienc.

\textbf{\textit{Impact} of Sampling temperature $\tau$ and LLM generation temperature $\alpha$} As shown in Figure \ref{fig:humaneval temperature analysis} when $\alpha$ remains constant, all values of $\tau$ enhance the diversity compared to vanilla generation, as opposed to search-based methods that often sacrifice diversity. Higher $\tau$ values increase diversity but can degrade performance by causing myopia. $\alpha$ significantly influences performance outcomes, with the best balance between diversity and performance achieved at moderate $\alpha$ values (0.3-0.6).


\textbf{More analysis} More analysis including agent world model and prompt sensitivity are in Appendix~\ref{sec: more anlaysis}.


\subsection{Planning as Reward Optimization\label{sec: reward optimization experiment}}

Recall that \methodname{} samples from a distribution to solve the trajectory optimization problem using LLM evaluation as the proxy for calculating future return. Similarly we can define such distribution for any target.
The same sampling technique could maximize the objective $\mathcal{J}$ left to right, achieving global-optimal planning. We examine how \methodname{} performs with two different types of objective:

\input{tables/agent_function}

\input{tables/orm}

\textbf{Guiding Agent Planning with Heuristics} 
We employ human-designed reward heuristics $J$ to guide LLM planning, using a semantic matching score to compare the current observation to the goal state, following the approach recommended by \citet{ma2024agentboard}. The heuristic rewards are sparse, requiring multiple steps to increase rewards. As shown in Table \ref{tab: agent with rm}, even without foresight, the reward enhances LLM agent performance. \methodname{} further overcomes the sparsity with foresight and improves the utility of reward functions.

\begin{wrapfigure}{r}{0.45\textwidth}  
    \centering  
    \includegraphics[width=\linewidth]{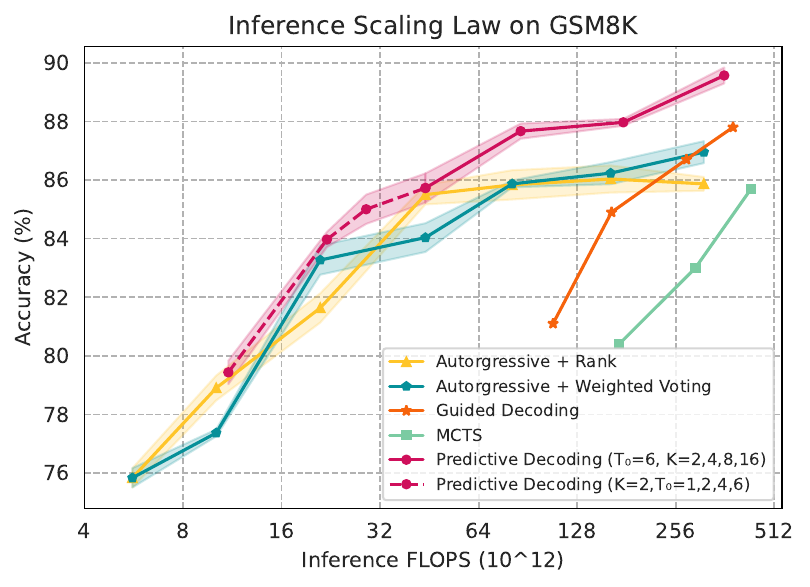}
    \caption{Inference compute scaling law on 189 examples sampled from GSM8K. All algorithms use Llama3-8B and Math-Shepherd. }  
    \label{fig:scaling law}  
\end{wrapfigure} 

\textbf{Guiding LLM Reasoning with Reward Model} We use Math-Shepherd~\citep{wang2023math} as the reward model for GSM8K. We compare \methodname{} against other reward model based planning methods, including: (1) Reward Model weighted Self-Consistency; (2) Reward Model based Ranking, selecting the top result from multiple samples; (3) Guided-Decoding; and (4) Monte Carlo Tree Search.
Results in Table \ref{tab: gsm8k with rm} demonstrate that all algorithms improve by over 10\% with Math-Shepherd. Although strong baselines like Guided-Decoding and MCTS are effective, they require extensive computation. \methodname{} achieves a 1.5\% improvement over Guided-Decoding with only 65\% of the computation and a 2.4\% improvement over MCTS with just 50\% of the computation. We further compare the sample efficiency of these methods in \S\ref{sec: inference scaling law}.

\subsection{Discussion: Inference Scaling Law\label{sec: inference scaling law}}

\citet{snell2024scaling} recently stressed the importance of inference scaling law, which states that LLM inference could benefit from using more computation to support more expressive algorithms. \citet{wu2024empirical} further discuss the efficiency-performance trade-off of different search algorithms. However, it is yet to be discussed if searching is the most efficient and effective choice for LLM reasoning.

We follow setting in \S\ref{sec: reward optimization experiment} using reward model Math-Shepherd. In Figure \ref{fig:scaling law}, we plot the inference scaling law of various methods and observe that: (i) Sampling-based methods have better computation efficiency compared to search-based methods. However autoregressive generation using simple reward ranking tends to saturate when more computation is available. (ii) \methodname{} which directly samples from optimal solution space achieves better scaling law than all sampling and search baselines. \methodname{} performance also consistently improves with more computation.





%% file: tables/reasoning.tex
\begin{table*}[!t]
\centering
\caption{The accuracy(\%) on GSM8K(eight-shot) and MATH (four-shot).\textsuperscript{\textdagger} \textsc{Codex} results are taken from \citet{xie2024self}. Inference FLOPS scale with the number of generated tokens, denoted as times of $k$ for proprietary models and calculated following ~\citep{kaplan2020scaling} for opensource models.}
\vspace{-0.3cm}
\begin{center}
\resizebox{0.98\linewidth}{!}{ 
\begin{tabular}{lrl|rc|c|c}
\toprule
Model & Size & Method & Inference FLOPS&\#Sample &GSM8K & MATH\\

\midrule
\midrule
\multirow{1.5}{*}{\textsc{Codex}} & \multirow{1.5}{*}{175B} & Autoregressive (PAL, \citet{gao2023pal})\textsuperscript{\textdagger} &  $k$ & $N=1$ & 72.0 & - 
\\
& & Guided-Decoding\citep{xie2024self}\textsuperscript{\textdagger} & $138.5k$ & $N=1$ & 80.2 & - 
\\
\midrule
\multirow{7.5}{*}{\textsc{Llama}-3} &  \multirow{7.5}{*}{8B}
& Autoregressive (PAL)& $7.5\times 10^{12}$ &$N = 1$& 71.3 & 29.7\\

& & Beam Search & $478.6\times 10^{12}$  & $N=1$ &  78.4 &  \textbf{34.4}    \\
& & Autoregressive + Self-Consistency& $59.8\times 10^{12}$ &\cellcolor{multisample}$N = 8$ & \cellcolor{multisample}80.9 & \cellcolor{multisample}31.8 \\
& & Beam Search + Self-Consistency & $478.6\times 10^{12}$ &\cellcolor{multisample}$N = 8$ & \cellcolor{multisample}79.5 & \cellcolor{multisample}39.4 \\
& & Guided-Decoding & $119.3\times 10^{12}$  &\cellcolor{multisample}$N = 8$& \cellcolor{multisample}63.9 & \cellcolor{multisample} -  \\
\cmidrule{3-7}
& & \methodname{} &$161.3\times 10^{12}$  &$N = 1$& \textbf{78.5} & 34.0\\
& & \methodname{} + Self-Consistency& $129.0\times 10^{13}$ &\cellcolor{multisample}$N = 8$& \cellcolor{multisample}\textbf{81.3} & \cellcolor{multisample}\textbf{40.3}\\
\midrule
\multirow{3}{*}{\textsc{Mistral-0.3}} & \multirow{3}{*}{7B}
& Autoregressive (PAL)& $6.6\times 10^{12}$ &$N = 1$ & 53.4 & 12.7 \\
& & Beam Search &  $418.8\times 10^{12}$  & $N = 1$ & 65.5 & 18.2 \\
\cmidrule{3-7}
& & \methodname{} & $141.1\times 10^{12}$ &$N = 1$ & \textbf{66.7} & \textbf{19.5}\\
\bottomrule
\end{tabular}}  
\end{center}
\label{tab: math main results}
\vspace{-0.5cm}
\end{table*}

%% file: tables/code.tex
\begin{table*}[!t]
\caption{The pass@1(\%) and pass@10(\%) rates on HumanEval (zero-shot) and MBPP (zero-shot). MBPP uses the 500 samples test set from huggingface.
\textsuperscript{\textdagger}Results are taken from \citet{roziere2023code}.} 
\vspace{-0.3cm}
\begin{center}
\resizebox{0.98\linewidth}{!}{ 
\begin{tabular}{lrl|r|cc|cc}
\toprule
\multirow{2}{*}{Model} & \multirow{2}{*}{Size} & \multirow{2}{*}{Method} & \multirow{1}{*}{FLOPS} &\multicolumn{2}{c|}{\humaneval} & \multicolumn{2}{c}{MBPP}\\
 & & & \multicolumn{1}{r|}{pass@1} & pass@1 & \cellcolor{multisample}pass@10 &  pass@1 & \cellcolor{multisample}pass@10 \\
\midrule
\midrule
\textsc{GPT-3.5 Turbo}\textsuperscript{\textdagger} & - &Autoregressive & - & 48.1 & \cellcolor{multisample}-  & 52.2 &\cellcolor{multisample}-  \\
\textsc{GPT-4}\textsuperscript{\textdagger} & - &Autoregressive & - &67.0 & \cellcolor{multisample}-  & - & \cellcolor{multisample}- \\
\midrule %
\multirow{4.5}{*}{\textsc{Llama}-3} & \multirow{4.5}{*}{8B}
& Autoregressive & $2.9\times 10^{12}$ &44.3 &\cellcolor{multisample}72.1 & 33.5 & \cellcolor{multisample}48.4 \\
& & Beam Search & $290.0\times 10^{12}$ & 53.5 & \cellcolor{multisample}70.1 & 33.6 & \cellcolor{multisample}50.2 \\
& & Self-infilling~\citep{zheng2023self} & $5.8 \times 10^{12}$&51.8 & \cellcolor{multisample}75.6  & 33.4 & \cellcolor{multisample}48.6\\
\cmidrule{3-8}
& & \methodname{} & $76.6\times 10^{12}$ &\textbf{57.3} & \cellcolor{multisample}\textbf{76.8} & \textbf{38.4} & \cellcolor{multisample}\textbf{52.6}\\
\midrule %
\multirow{4.5}{*}{\textsc{Deepseek~Coder}} & \multirow{4.5}{*}{6.7B}
& Autoregressive & $2.4\times 10^{12}$ & 45.2 & \cellcolor{multisample}66.5 & 38.6 & \cellcolor{multisample}53.8 \\
& & Beam Search & $240.0\times 10^{12}$ & 46.7 & \cellcolor{multisample}50.0 & \textbf{39.7} & \cellcolor{multisample}42.6\\
& & Self-infilling & $4.8\times 10^{12}$ & 46.3 & \cellcolor{multisample}76.2 & 38.4 &  \cellcolor{multisample}51.6 \\
\cmidrule{3-8}
& & \methodname{} & $62.4\times 10^{12}$ & \textbf{47.6} & \cellcolor{multisample}\textbf{79.9} & 
34.0 & \cellcolor{multisample}\textbf{55.2} \\
\bottomrule
\end{tabular}}  
\end{center}
\label{tab: coding main results}
\vspace{-0.5cm}
\end{table*}

%% file: tables/agent.tex
\begin{table*}[!t]
\caption{The success rates(\%) and the progress rates(\%) on AlfWorld (one-shot) and PDDL (one-shot) within 20 steps. \methodname{} uses the same prompt as Act. }
\label{table: code}
\vspace{-0.3cm}
\begin{center}
\resizebox{0.98\linewidth}{!}{ 
\begin{tabular}{lrl|rc|cc|cc}
\toprule
\multirow{2}{*}{Model} & \multirow{2}{*}{Size} & \multirow{2}{*}{Method}&  \multirow{2}{*}{FLOPS}&\multirow{2}{*}{\# Sample} & \multicolumn{2}{c|}{\alfworld} & \multicolumn{2}{c}{PDDL}\\
 & & & &  & Success & Progress & Success & Progress\\
\midrule
\midrule %
\multirow{3}{*}{\textsc{GPT-3.5 Turbo}} & \multirow{3}{*}{-}
& Act~\citep{yao2022react} & $k$ &$N=1$ &6.7 & 25.6 &  5.0& 20.0\\
& & ReAct~\citep{yao2022react} & $8.4k$ &$N=1$ &7.5 & \textbf{34.8} & 3.3 & 15.6\\
\cmidrule{3-9}
& & \methodname{} & $40.7k$& $N=1$  &\textbf{14.9} & 33.2 & \textbf{11.7} & \textbf{28.9}\\ 
\midrule %
\multirow{3.5}{*}{\textsc{Llama}-3.1} & \multirow{3.5}{*}{70B}
& Act & $43.2\times 10^{12}$ &$N=1$ &  19.4 & 43.0 & 35.0 & 54.6\\
& & ReAct & $488.0\times 10^{12}$ &$N=1$ &   18.7 & 43.4 & 18.3 & 36.9\\
\cmidrule{3-9}
& & \methodname{} & $224.1\times 10^{13}$   & $N=1$&\textbf{44.0} & \textbf{53.9} & \textbf{38.3} & 
\textbf{59.7}\\
\bottomrule
\end{tabular}}  
\end{center}
\label{tab: agent main results}
\vspace{-0.3cm}
\end{table*}

%% file: tables/ablation.tex
\begin{table*}[!t]
\caption{Impact of Foresight $T_0$, sampling number $K$ and trajectory recycle \includegraphics[height=10pt]{figures/recycle_icon3.png}. FLOPS of autoregressive generation is denoted as $k$ and is scaled based on number of generated tokens.}
\vspace{-0.3cm}
\centering
\resizebox{0.98\linewidth}{!}{ 
\begin{tabular}{c|ccc|c|ccc}
\toprule
  \# Foresight $T_0$  & \multicolumn{1}{c}{\textsc{FLOPS}} & \gsm & \humaneval &   \# Sampling $K$    & \multicolumn{1}{c}{\textsc{FLOPS}}  & \humaneval\ w. \includegraphics[height=10pt]{figures/recycle_icon3.png}, &  wo.\includegraphics[height=10pt]{figures/recycle_icon3.png}\\ \midrule\midrule
$T_0=1$ &       $3.6k$          &   77.0    &  45.1         & $K = 2$ &     $6.6k$             &  51.2     &  47.6         \\
$T_0=2$ &       $7.2k$           &  76.4     &     48.8      & $K = 4$ &    $13.2k$             &    54.3   &   46.3        \\
$T_0=4$&       $14.4k$           &   77.9    &     52.4      & $K = 6$ &        $19.9k$          &        50.6     &     49.4 \\
$T_0=6$ &     $21.5k$ &    78.5        &  57.3     & $K = 8$ &    $26.4k$      &   57.3    &   49.4    \\ 
\bottomrule
\end{tabular}
}
\label{tab: ablation study}
\vspace{-0.3cm}
\end{table*}

%% file: tables/agent_function.tex
\begin{table*}[!t]
\caption{The success rates(\%) and the progress rates(\%) on AlfWorld (one-shot) and PDDL (one-shot) with different large language models and matching function as heuristics.}
\vspace{-0.3cm}
\begin{center}
\resizebox{0.98\linewidth}{!}{ 
\begin{tabular}{llc|cc|cc}
\toprule
\multirow{2}{*}{Model} &  \multirow{2}{*}{Method}& \multirow{2}{*}{\# Foresight $T_0$} & \multicolumn{2}{c|}{\alfworld} & \multicolumn{2}{c}{PDDL}\\
 & & & Success & Progress & Success & Progress\\
\midrule
\midrule %
\multirow{3.5}{*}{\textsc{GPT-3.5 Turbo}}
& Act~\citep{yao2022react} &  $T_0=0$ & 6.7& 25.6 &  5.0 &20.0\\
& ReAct~\citep{yao2022react} & $T_0=0$ & 7.5 & 34.8 & 3.3 & 15.6\\
\cmidrule{2-7}
&\methodname{} + Reward & $T_0=0$ & 7.5 & 29.5 & 6.7& 23.9  \\
&\methodname{} + Reward & $T_0=6$ & \textbf{13.4} & \textbf{37.6} & \textbf{8.3} & \textbf{27.9} \\
\bottomrule
\end{tabular}}  
\end{center}
\label{tab: agent with rm}
\vspace{-0.4cm}
\end{table*}

%% file: tables/orm.tex
\begin{table*}[!t]
\centering
\caption{The accuracy(\%) on GSM8K (8-shot) with reward model(RM) Math-Shepherd.}
\label{tab: gsm8k with rm}
\vspace{-0.3cm}
\begin{center}
\resizebox{0.98\linewidth}{!}{ 
\begin{tabular}{lrp{5cm}rr|c}
\toprule
Model & Size & Method & Inference FLOPS & Reward FLOPS &GSM8K \\

\midrule
\midrule
\multirow{7}{*}{\textsc{Llama}-3} & \multirow{7}{*}{8B}
& Autoregressive & $5.6\times 10^{12}$ &  $0.0\times 10^{12}$ & 70.4 \\
&  & RM-weighted Self-Consistency & $44.5\times 10^{12}$ & $0.4\times 10^{12}$ & 82.8 \\
& & RM-based Ranking  & $44.5\times 10^{12}$  & $0.4\times 10^{12}$ &85.9 \\
& & Guided-Decoding + RM & $276.2\times 10^{12}$  & $11.4\times 10^{12}$ &86.5 \\
& & Monte Carlo Tree Search + RM & \cellcolor{multisample}$360.0\times 10^{12}$ & \cellcolor{multisample}$16.2\times 10^{12}$ & \cellcolor{multisample}85.5\\

\cmidrule{3-6}
& & \methodname{} + RM & $182.3\times 10^{12}$ & $2.9\times 10^{12}$
 & 87.9  \\
 
& & \methodname{} + RM & \cellcolor{multisample}$360.5\times 10^{12}$ & \cellcolor{multisample}$5.6\times 10^{12}$
 & \cellcolor{multisample}\textbf{89.9}  \\

\bottomrule
\end{tabular}}  
\end{center}
\vspace{-0.5cm}
\end{table*}

%% file: paragraphs/07_related_work.tex
\textbf{LLM-based Planning and Reasoning} One major development is the emergence of LLM step-by-step reasoning abilities \citep{wei2022chain}. This capability is further enhanced by more expressive inference algorithms like searching~\citep{yao2022react, yao2024tree, hao2023reasoning, wang2023voyager, zheng2023self, xie2024self, zhao2024empowering,sun2024adaplanner}.
However, most of these algorithms are computation-intensive. Recent work discusses the inference time scaling law for LLM reasoning \citep{wu2024empirical, snell2024scaling}. This highlights the need for an efficient yet global-optimal method. Our work attempts to close this research gap by introducing \methodname{}. Another line of work studies the expressiveness of LLM reasoning. \citet{feng2024towards} confirms LLM reasoning's expressiveness equals dynamic programming. \citet{li2024chain} discusses Turing completeness in chain-of-thought reasoning. Our theory does not contradict their claims but examines whether any pretrained LLM could find such an optimal solution.

\paragraph{Model Predictive Control}
Model Predictive Control (MPC) is a widely used control strategy that involves solving an optimization problem at each time step by forecasting future results \citep{witkin1988spacetime, qin1997overview}. Similarly, in model-based reinforcement learning (RL), an agent aims to maximize future success by using a dynamics model to simulate state transitions \citep{silver2017mastering, ha2018world, anthony2017thinking, racaniere2017imagination, nagabandi2018neural}. These algorithms enable agents to think ahead, envisioning the outcomes of various potential actions, and making deliberate selections among alternatives. Our method follows this line of work.


\paragraph{Combinatorial Optimization via Sequential Sampling} Sampling-based methods have been extensively used to solve combinatorial optimization efficiently \citep{sun2023difusco, janner2022planning, qin2022cold, du2024learning}, by constructing an energy-based model where the generative probability reflects the objective. However, these methods often use non-autoregressive sampling, which undermines the Markov property of sequences critical to tasks like planning and language modeling. Autoregressive diffusion models \citep{wu2023ar, chen2024diffusion} have been proposed to sample based on optimization constraints while enabling causal generation. However, these methods rely on diffusion training loss and cannot be directly applied to LLMs. In this work, we follow \citet{deng2020residual, ji2023language} and use the sampling-importance-resampling (SIR) technique, which enables us to maintain autoregressive next-token prediction while achieving global optimality. This method can be easily combined with model predictive control and used on any LLM.

%% file: paragraphs/final_parts.tex
\section{Conclusions and Limitation}
In this paper, we analyzed the limitations and potential of Large Language Models (LLMs) in planning and reasoning tasks, particularly focusing on their myopic tendencies. We introduced \methodname{}, which employs Model Predictive Control to enhance non-myopic planning, significantly improving performance across various domains. Our experiments confirmed \methodname{}'s effectiveness in boosting planning accuracy and computational efficiency. These results open promising avenues for future research into optimizing LLM reasoning and incorporating foresight into decision-making, paving the way for more robust and efficient LLM applications.


The main limitation of our work is that we are targeted towards improving reasoning and planning at inference times, without exploring how our method could aid in generating training data—an advantage of search-based methods. Additionally, \methodname{} is limited to tasks with clear reasoning steps, like math or coding. While it potentially could aid global awareness in free-form generation, such as long-context tasks, we leave this for future research.

%% file: paragraphs/appendix.tex
\newpage
\section{Pseudo Code\label{sec: pseudocode}}
\input{tables/algorithm}

\section{Theoretical Results}

\subsection{Sample Efficiency of Model Predictive Control \label{sec: sample efficiency}}

In this section, we explain in detail why model predictive control has the potential to improve sample efficiency in solving trajectory optimization problems. 

We begin by introducing a stochastic setting, where trajectory $\tau = (a_0,a_1,\dots, a_T)$ is a chain of actions that could take values in set $\mathcal{A}$. A policy model models the distribution of trajectories as $\pi_{\theta}(\tau)$, denoting the probability that the policy model samples the trajectory. 

For the best scenario, the policy model is perfect and the sample $\tau' \sim \pi_{\theta}(\tau)$ maximizes the objective $\mathcal{J}(\tau)$ and $\tau'$ is the best solution. However, in most cases our LLM is not the perfect policy model and prone to make errors during sampling. Instead, we sample multiple solutions $\tau_1,\tau_2,\dots,\tau_k \sim \pi_{\theta}(\tau)$ and selects the best solution $\tau' = \max_{i=1,\ldots,k}\mathcal{J}(\tau_i)$. A different method is using model predictive control during sampling. We thereby discuss the number of $k$ necessary to obtain the correct $\tau'$.

\begin{proposition}Given the optimal trajectory $\tau'=(a_0',\dots,a_T')$ that maximizes $\mathcal{J}$. $\tau^1,\tau^2,\dots,\tau^{k|\mathcal{A}|}\sim \pi_{\theta}(\tau)$. The best $a_0^r$ is the first step of the trajectory $\tau^j$ that maximizes $\mathcal{J}$.

If $a_0^m$ satisfies:
\begin{equation}
    a_0^m = \underset{a_0}{\arg\max}\left[\max_{i=1\dots k}\mathcal{J}(a_0,\tau_{\rightarrow a_0}^i)\right], \tau_{\rightarrow a_0}^i\sim \pi_{\theta}(\cdot\mid a_0)
\end{equation}
Then $P(a_0^m=a_0')\geq P(a_0^j=a_0')$
\end{proposition}

\textit{Informal proof}: If $\pi$ has no knowledge of $\mathcal{J}$, then the best case in sampling would be $\tau^1,\tau^2,\dots,\tau^{k|\mathcal{A}|}$ first sample $a_0$ from $U(a_0)$, then samples $\tau_{\rightarrow a_0}\sim\pi_{\theta}(\cdot\mid a_0)$, in order to gain thorough information about which $a_0$ would be the best for the trajectory. In this case, the sampling process is same as model predictive control and thus the best case results in $P(a_0^m=a_0') = P(a_0^j=a_0')$. However, when $\pi_{\theta}$ places more importance on $a_0\not=a_0'$, i.e. $\exists a_0, \pi(a_0^*)>\pi(a_0')$, there exists a larger possibility that $a_0^*$ is selected by $J$ instead of $a_0'$ than model predictive control, as let $n$ be the times $a_0^*$ is sampled, $k>n$, then $E\left[\max_{i=1,\dots,n}\mathcal{J}(a_0^*, \tau_{\rightarrow a_0^*}^i)\right]\geq E\left[\max_{i=1,\dots,k}\mathcal{J}(a_0^*, \tau_{\rightarrow a_0^*}^i)\right]$. Thus the probability that $a_0^*$ is selected would be larger and $P(a_0^m=a_0') \geq P(a_0^j=a_0')$. 

This proposition show that if directly sample solution from an exponential solution space, it would be unlikely to obtain the optimal solution. We also showcase this phenomenon that model predictive control improves sample efficiency than direct sampling through simulated experiments.

\begin{figure}[htbp]

    \begin{minipage}{0.49\textwidth} 
    \centering
 \includegraphics[width=0.7\linewidth]{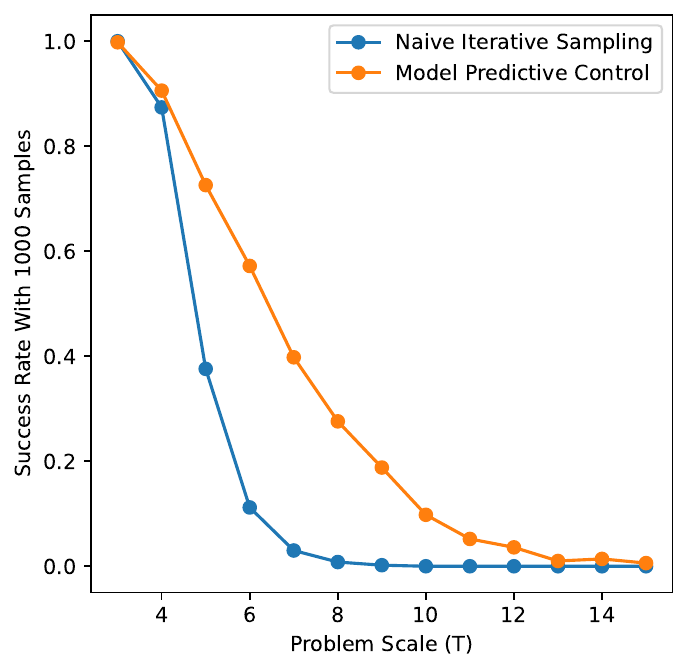}
\caption{Pass@1000 accuracy v.s. problem scale on the simulation task. $\pi_{\theta}$ follows a uniform distribution. The performance of iterative sampling decreases significantly when the problem scales up.}
    \label{fig:simulation experiment scale}
    \end{minipage}
  \hfill
    \begin{minipage}{0.49\textwidth} 
    \centering
\includegraphics[width=0.7\linewidth]{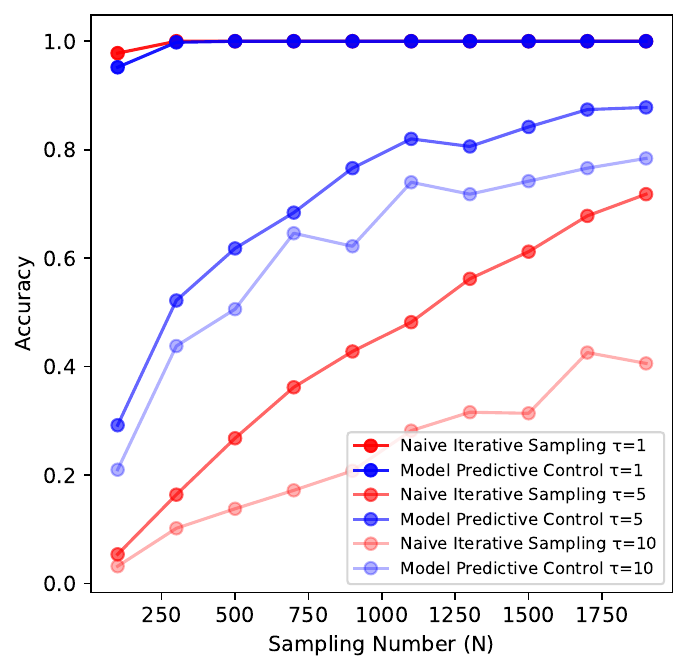}
    \caption{Pass@k accuracy on simulation task with different sampling numbers. Larger $\tau$ denotes that the sampling distribution $\pi_{\theta}$ is close to ground truth model.}
    \label{fig:simulation experiment tau}
    \end{minipage}
\end{figure}

We focus on a very simple combinatorial optimization problem, where a list $(a_0,a_1,\dots,a_T)$ satisfies $a_0 \in \mathcal{A}$ takes the value from $|\mathcal{A}|$ integers. Our objective is to find the list of numbers with maximum sum. We control the sample complexity of 
``random sampling + ranking'' and ``model predictive control''. Results are shown in Figure \ref{fig:simulation experiment scale} and Figure \ref{fig:simulation experiment tau}. On all settings, MPC is more sample efficient than naive iterative sampling. Notably, we alter the prior distribution $\pi_{\theta}$ such that it samples the number based on value $e^{\text{number}/\tau}$. Larger $\tau$ is closer to uniform distribution yet smaller $\tau$ is closer to the optimal distribution. We can see that a good prior always results in improved performances, and the gap between iterative sampling and MPC is smaller. However, when the $\pi_{\theta}$ is not good (e.g. using a LLM that is not well-pretrained), the sample efficiency of MPC is much better than iterative sampling.


    

\subsection{Proof for Proposition \ref{thm: main}\label{sec: proof 4.1}}

    The distribution that solves the optimization problem in Eq.\ref{eq: solution} is in the form of: 
    \begin{equation}
        p_{\tau}(a_t)\propto P^{\text{LLM}}\left(a_t\mid \mathbf{a}_{<t}^{\prime}, \mathbf{s}_{<t}^{\prime}\right)\exp\left[\mathbb{E}_{\mathbf{a}_{>t},\mathbf{s}_{>t}} P^{\text{LLM}}\left(a_t, \mathbf{a}_{>t}, \mathbf{s}_{>t} \mid \mathbf{a}_{<t}^{\prime}, \mathbf{s}_{<t}^{\prime}\right)/\tau\right].
    \end{equation}

\begin{proof}
    The goal is to solve this optimization problem: 
    $$
    a_t^{\prime}=\underset{a_t}{\arg \max}\ \mathbb{E}_{\mathbf{a}_{>t}, \mathbf{s}_{>t}} P^{\text{LLM}}\left(a_t, \mathbf{a}_{>t}, \mathbf{s}_{>t} \mid \mathbf{a}_{<t}^{\prime}, \mathbf{s}_{<t}^{\prime}\right).
    $$
    Energy-based Model can transfer the optimization problem into probability distribution, such as,

    \begin{lemma}\citep{lecun2006tutorial} Let $\vf=\{f_i\}_{i=1\dots n}$ is a set of evaluation metrics, the distribution $p(x)$ solves 
    the optimization problem $x = \underset{x\in\mathcal{P}}{\arg\min}\sum_i\mu_if_i(x)$, only if $p(x)$ satisfies: 
    
    \begin{equation}\label{eq: ebm}
        p(x) \propto p_{\theta}(x)\exp\left[-E_\mu(x)\right], \forall x \in \mathcal{P}
    \end{equation}

    where $\mathcal{P}$ is the support set of probability distribution of $p_{\theta}$. 
    \end{lemma}

Here $a_t$ is generated by LLM through autoregressive generation, i.e. $a_t\sim P^{\text{LLM}}(a_t\mid \mathbf{a}_{<t},\mathbf{s}_{<t})$, and the constraint can be rewritten as $f(a_t) = \mathbb{E}_{\mathbf{a}_{>t}, \mathbf{s}_{>t}} P^{\text{LLM}}\left(a_t, \mathbf{a}_{>t}, \mathbf{s}_{>t} \mid \mathbf{a}_{<t}^{\prime}, \mathbf{s}_{<t}^{\prime}\right) $. Plugging into Eq. \ref{eq: ebm} and we would yield $p{a_t}$. $\tau$ can be added as a temperature coefficient and does not lead to difference in optimization. 
\end{proof}

\section{More Analysis\label{sec: more anlaysis}}

\begin{figure}[t!]  
\vspace{-5mm}
  \centering   
  \begin{minipage}{0.3\textwidth}  
    \includegraphics[width=\textwidth]{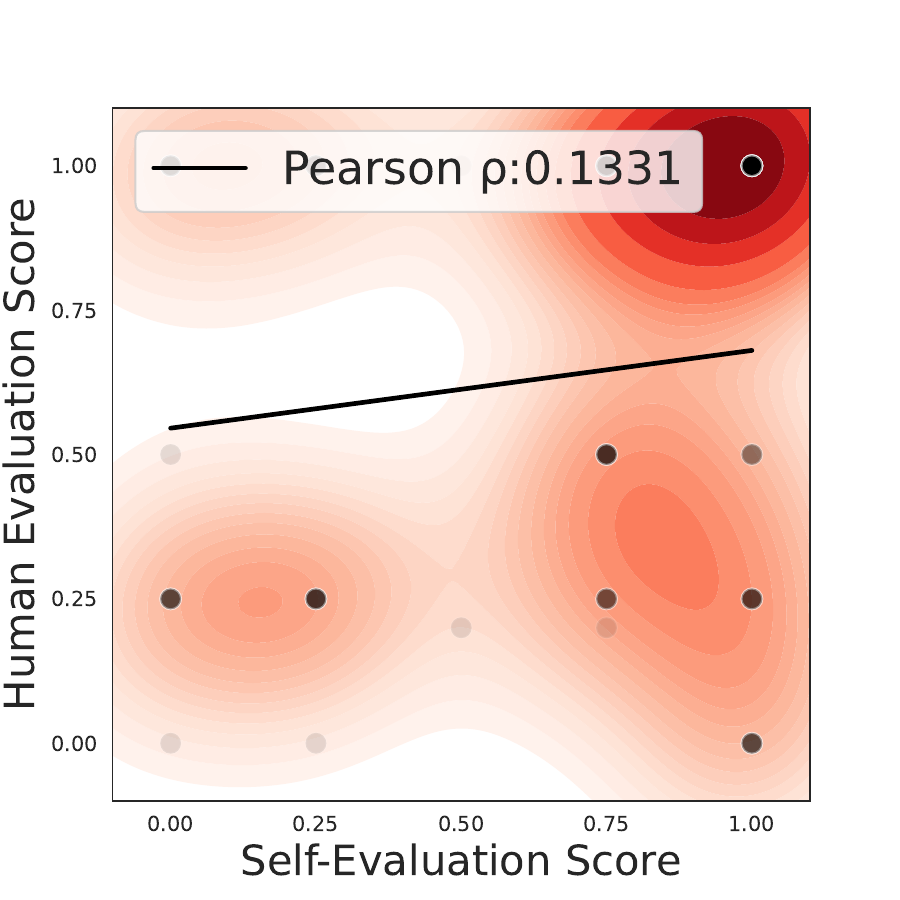}
  \end{minipage}  
  \begin{minipage}{0.3\textwidth}  
    \includegraphics[width=\textwidth]{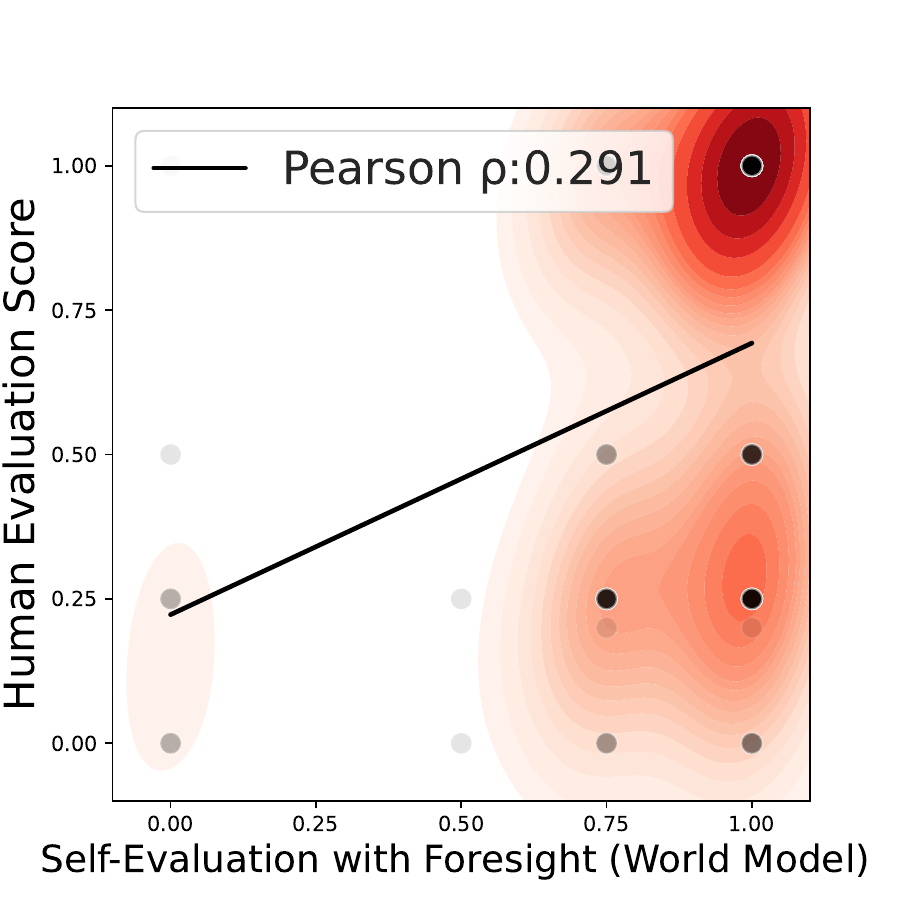}
  \end{minipage} 
  \begin{minipage}{0.3\textwidth}  
    \includegraphics[width=\textwidth]{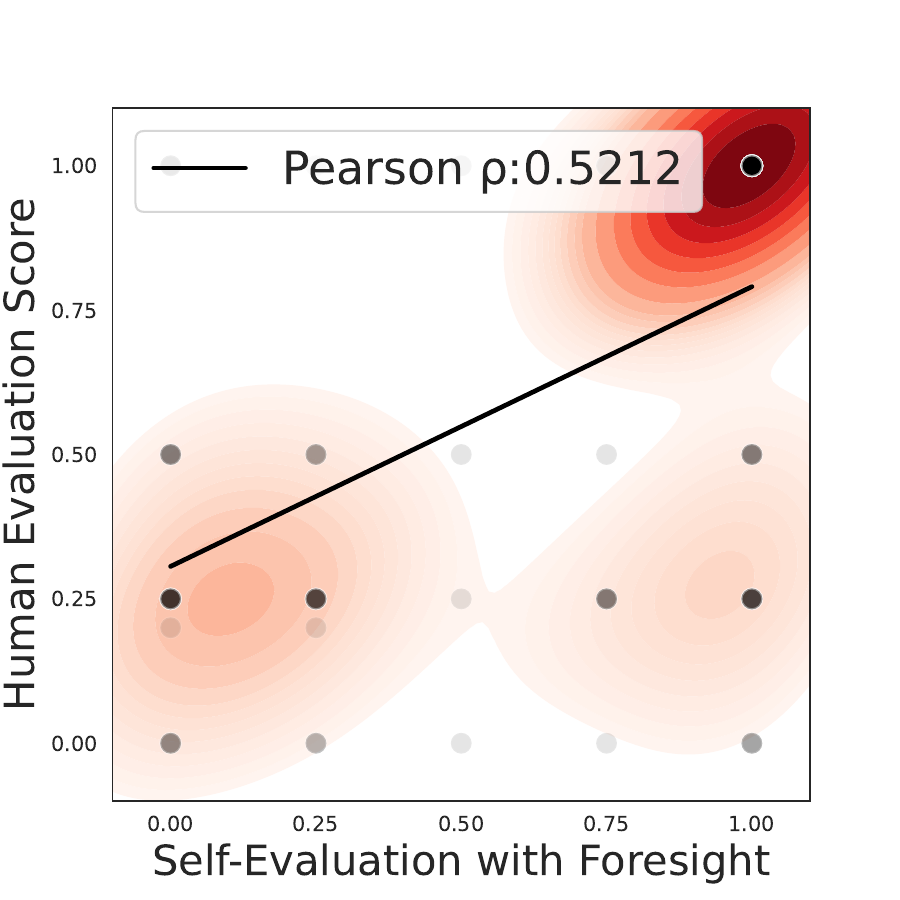}
  \end{minipage} 
  \caption{Comparison of GPT-3.5-Turbo Self-Evaluation $\hat{r}(s,a)$ w/o. modeling on 152 trajectory data from AlfWorld. We compare from left-to-right (1) LLM direct self-evaluation based on past trajectory($\rho=0.133$),(2) LLM first perform model rollout and then perform self-evaluation based on both past and future trajectory($\rho=0.291$), (3)substituting LLM-as-world-model for an accurate world model for rollout($\rho=0.521$) against human evaluation. All scores are chosen from \{0,0.25,0.5,0.75,1\} and we use KDE plots to visualize overlapping discrete points.}
  \label{fig: agent eval calibration}
\end{figure}

\paragraph{\textit{Impact} of Agent World Model Hallucination} \methodname{} requires the LLM to act as a world model to infer states after each step. However, we often observe hallucinations. For instance, in AlfWorld, when the agent tries to find an apple and place it on desk 1, the imagined state after "open fridge 1" might incorrectly include an apple in the fridge, even if it's not there. This reflects an inherent model bias issue~\citep{kurutach2018model}, where the world model fails to accurately represent the environment. Figure \ref{fig: agent eval calibration} shows that this results in lower LLM score accuracy compared to using real environment transitions. Most evaluation errors are false positives, indicating LLM overconfidence in predicting step outcomes. Increasing $\tau$ to 0.05-0.1 helps mitigate this by preventing overfitting.

\begin{figure}
    \centering
    \begin{minipage}{0.5\textwidth} 
    \includegraphics[width=\linewidth]{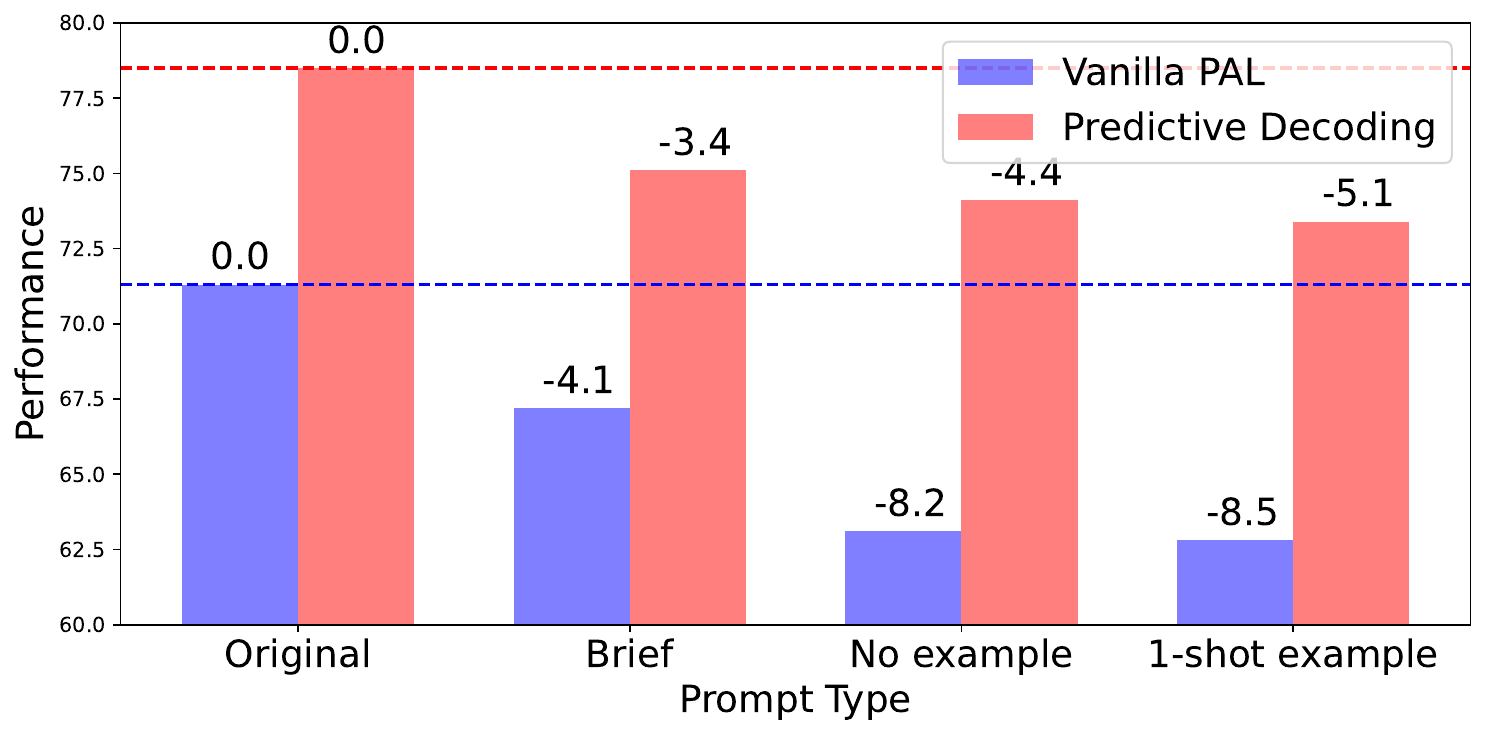}
    \caption{Prompt sensitivity to instruction brevity, as well as the number of in context samples on GSM8K. Both methods use Llama3-8B.}
    \label{fig:prompt sensitivity}
    \end{minipage}
\end{figure}

\paragraph{Prompt Sensitivity Analysis} 
\citet{wang2024chain} proposes an intereting problem: improving LLM reasoning without prompting. We find \methodname{} can also improves performances and maintain high performance even without in-context samples and detailed instructions. As shown in Figure \ref{fig:prompt sensitivity}, vanilla generation with LLM show significant drop in performance when the detailed instructionsa are not given, or when there are no in-context examples. \methodname{} shows smaller decline in performance, demonstrating better prompt sensitivity. Moreover, \methodname{} without in-context examples could outperforms vanilla generation with in-context examples. This demonstrate our method also has the potential to improve reasoning without prompting.

\section{Further Discussion on Related Work}
\paragraph{Combinatorial Optimization via Sequential Sampling} Sampling-based methods have been extensively used to solve combinatorial optimization efficiently \citep{sun2023difusco, janner2022planning, qin2022cold, du2024learning}, by constructing an energy-based model where the generative probability reflects the objective. However, these methods often use non-autoregressive sampling, which undermines the Markov property of sequences critical to tasks like planning and language modeling. Autoregressive diffusion models \citep{wu2023ar, chen2024diffusion} have been proposed to sample based on optimization constraints while enabling causal generation. However, these methods rely on diffusion training loss and cannot be directly applied to LLMs. In this work, we follow \citet{deng2020residual, ji2023language} and use the sampling-importance-resampling (SIR) technique, which enables us to maintain autoregressive next-token prediction while achieving global optimality. This method can be easily combined with model predictive control and used on any LLM.

\section{Case Study~\label{sec: case study}}

\paragraph{Case Study for Reasoning} Here we provide the generations to show how \methodname{} solves a math problem.

\begin{mdframed}
    Question: Toulouse has twice as many sheep as Charleston. Charleston has 4 times as many sheep as Seattle. How many sheep do Toulouse, Charleston, and Seattle have together if Seattle has 20 sheep?

    def solution():

    \textcolor{brown}{\{`    \#Toulouse has twice as many sheep as Charleston. Charleston has 4 times as many sheep as Seattle. How many sheep do Toulouse, Charleston, and Seattle have together if Seattle has 20 sheep?': 1.0\}}

    Line 1: \#Toulouse has twice as many sheep as Charleston. Charleston has 4 times as many sheep as Seattle. How many sheep do Toulouse, Charleston, and Seattle have together if Seattle has 20 sheep?

    \textcolor{brown}{\{`        seattle\_sheep = 20': 1.0\}}

    Line 2: seattle\_sheep = 20

   \textcolor{brown}{ \{`    charleston\_sheep = seattle\_sheep * 4': 0.9767, `    charleston\_sheep = 4 * seattle\_sheep': 0.0233\}}

    Line 3: charleston\_sheep = seattle\_sheep * 4

   \textcolor{brown}{ \{`    toulouse\_sheep = charleston\_sheep * 2': 0.9377, `    toulouse\_sheep = seattle\_sheep * 2': 0.0623\}}
    
    Line 4: toulouse\_sheep = charleston\_sheep * 2

    \textcolor{brown}{\{`    total\_sheep = seattle\_sheep + charleston\_sheep + toulouse\_sheep': 1.0\}}

     Line 5: total\_sheep = seattle\_sheep + charleston\_sheep + toulouse\_sheep'

    \textcolor{brown}{\{`    result = total\_sheep': 0.9999\}}

     Line 6: result = total\_sheep

    \textcolor{brown}{\{`    return result':1.0\}}
    
    Line 7: return result
     
\end{mdframed}

Examples in trajectory pool:

Foresight 1 for the first line (composed of 6 steps): 
\begin{lstlisting}[language=json,firstnumber=1]
{'Action': None, 'Reward': 0.9984305701520592}

{'Action': '    """Toulouse has twice as many sheep as Charleston. Charleston has 4 times as many sheep as Seattle. How many sheep do Toulouse, Charleston, and Seattle have together if Seattle has 20 sheep?"""\n', 'Reward': 0.9984305701520592}

{'Action': '    seattle_sheep = 20\n',  'Reward': 0.9984305701520592}

{'Action': '    charleston_sheep = seattle_sheep * 4\n', 'Reward': 0.9984305701520592}, 

{'Action': '    toulouse_sheep = charleston_sheep * 2\n', 'Reward': 0.9984305701520592}

{'Action': '    total_sheep = seattle_sheep + charleston_sheep + toulouse_sheep\n', 'Reward': 0.9984305701520592}
\end{lstlisting}

Foresight 1 for the third line (composed of 4 steps): 
\begin{lstlisting}[language=json,firstnumber=1]

{'Action': '    charleston_sheep = seattle_sheep * 4\n', 'Reward': 0.9842750955836792}, 

{'Action': '    toulouse_sheep = 2 * charleston_sheep\n',  'Reward': 0.9842750955836792}

{'Action': '    total_sheep = seattle_sheep + charleston_sheep + toulouse_sheep\n', 'Reward': 0.9842750955836792}

{'Action': '    result = total_sheep\n', 'Reward': 0.9842750955836792}

{'Action': '    return result\n',  'Reward': 0.9842750955836792}

\end{lstlisting}

From the case study we could observe that for easier steps, the distribution is concentrated on a single action. This is due to LLM generation is inherently confident on these lines. Our method mainly helps improving the lines where LLM generation is less confident (e.g. Line 3 \& 4).

\paragraph{Case Study for Agent} In this section we also provide a case study for agent task. Agent tasks are different from reasoning tasks in two ways: (i) Agent tasks make predictions not only based on prior actions, but also on environment observations. Therefore, during foresight, the agent needs to use LLM as a world model~\citep{hao2023reasoning} to infer the next state. (ii) Exploration is crucial for agents. Therefore maintaining stochastic during sampling is important. 

\begin{mdframed}
    Goal: look at bowl under the desklamp.

    Observation: You are in the middle of a room. Looking quickly around you, you see a bed 1, a desk 2, a desk 1, a drawer 6, a drawer 5, a drawer 4, a drawer 3, a drawer 2, a drawer 1, a garbagecan 1, a laundryhamper 1, a safe 1, a shelf 6, a shelf 5, a shelf 4, a shelf 3, a shelf 2, and a shelf 1.

    \textcolor{brown}{\{`go to desk 1': 0.9975, `go to desk 2': 0.0025\}}

    Step 1: go to desk 1

    Observation: use desklamp 1

    \textcolor{brown}{\{'use desklamp 1': 0.4530, 'examine desklamp 1': 0.5469, 'examine desk 2': 8.551e-06\}}

    Step 2: use desklamp 1

    Observation: You turn on the desklamp 1.

   \textcolor{brown}{ \{'examine desk 1': 0.9822, 'examine desklamp 1': 0.0177\}}

    Step 3: examine desk 1

    Observation: On the desk 1, you see a desklamp 1, a keychain 3, a mug 2, a pen 2, a pen 1, and a pencil 2.

   \textcolor{brown}{ \{'examine desklamp 1': 0.5711, 'go to desk 2': 0.4289\}}

   Step 4: examine desklamp 1

   Observation: Nothing happens.

    \textcolor{brown}{\{'go to shelf 1': 0.9974, 'go to desk 2': 0.00258\}}

     Step 5: examine shelf 1

     Observation: On the shelf 1, you see a cellphone 3, and a creditcard 2.

    \textcolor{brown}{\{'go to desk 2': 0.8146, 'go to shelf 2': 0.1853\}}

     Step 6: go to desk 2

     Observation: On the desk 2, you see a alarmclock 1, a bowl 1, and a cd 3.

    \textcolor{brown}{\{'take bowl 1 from desk 2': 0.9626, 'examine bowl 1': 0.03733\}}

     Step 7: take bowl 1 from desk 2

    Observation: You pick up the bowl 1 from the desk 2.
    
    \textcolor{brown}{\{'go to desk 1': 1.0\}}

    Step 8: go to desk 1

    Observation: On the desk 1, you see a desklamp 1, a keychain 3, a mug 2, a pen 2, a pen 1, and a pencil 2. The task is finished.  
\end{mdframed}

Examples in trajectory pool:

Foresight 1 for the first step (composed of 5 steps): 
\begin{lstlisting}[language=json,firstnumber=1]

{'Action': None, 'Observation': 'You are in the middle of a room. Looking quickly around you, you see a bed 1, a desk 2, a desk 1, a drawer 6, a drawer 5, a drawer 4, a drawer 3, a drawer 2, a drawer 1, a garbagecan 1, a laundryhamper 1, a safe 1, a shelf 6, a shelf 5, a shelf 4, a shelf 3, a shelf 2, and a shelf 1.', 'Reward': 0.9281983462766966}

{'Action': 'go to desk 1', 'Observation': 'On the desk 1, you see a book 1, a computer 1, a desklamp 1, a notebook 1, and a pen 1.', 'Reward': 0.9281983462766966}

{'Action': 'examine desklamp 1', 'Observation': 'The desklamp 1 is currently turned off.', 'Reward': 0.9281983462766966}

{'Action': 'use desklamp 1', 'Observation': 'You turn on the desklamp 1.', 'Reward': 0.9281983462766966}

{'Action': 'examine desk 1',  'Observation': 'On the desk 1, you see a book 1, a computer 1, a desklamp 1, a notebook 1, a pen 1, and nothing is under the desklamp 1.', 'Reward': 0.9281983462766966}

{'Action': 'go to desk 2', 'Observation': 'On the desk 2, you see a bowl 1, a cup 1, and a desklamp 2.', 'Reward': 0.9281983462766966}
\end{lstlisting}

Foresight 1 for the second step (composed of 5 steps): 

\begin{lstlisting}[language=json,firstnumber=1]

{'Action': 'go to desk 1','Observation': 'On the desk 1, you see a desklamp 1, a keychain 3, a mug 2, a pen 2, a pen 1, and a pencil 2.', 'Reward': 0.9279943292829683}

{'Action': 'use desklamp 1',  'Observation': 'You turn on the desklamp 1.', 'Reward': 0.9279943292829683}

{'Action': 'examine desk 1', 'Observation': 'On the desk 1, you see a desklamp 1, a keychain 3, a mug 2, a pen 2, a pen 1, a pencil 2, and a drawer 1.', 'Reward': 0.9279943292829683}

{'Action': 'open drawer 1', 'Observation': 'You open the drawer 1. Inside, you see a bowl 1.', 'Reward': 0.9279943292829683}

{'Action': 'take bowl 1 from drawer 1', 'Verified': None, 'Observation': 'You pick up the bowl 1 from the drawer 1.', 'Reward': 0.9279943292829683}

{'Action': 'put bowl 1 on desk 1', 'Observation': 'You put the bowl 1 on the desk 1, under the desklamp 1.', 'Reward': 0.9279943292829683}
\end{lstlisting}

Foresight 2 for the second step (composed of 5 steps, lower quality):
\begin{lstlisting}[language=json,firstnumber=1]

{'Action': 'go to desk 1', 'Observation': 'On the desk 1, you see a desklamp 1, a keychain 3, a mug 2, a pen 2, a pen 1, and a pencil 2.', 'Reward': 0.7375680280039042}

{'Action': 'examine desklamp 1',  'Observation': 'The desklamp 1 is currently turned off.', 'Reward': 0.7375680280039042}

{'Action': 'use desklamp 1', 'Observation': 'You turn on the desklamp 1.', 'Reward': 0.7375680280039042}

{'Action': 'go to desk 2', 'Observation': 'On the desk 2, you see a book 1, a bowl 1, a clock 1, and a stapler 1.', 'Reward': 0.7375680280039042}

{'Action': 'examine bowl 1', 'Observation': 'You are now looking at the bowl 1 under the light from the desklamp 1.', 'Reward': 0.7375680280039042}

{'Action': 'inventory',  'Observation': 'You have nothing in your inventory.', 'Reward': 0.7375680280039042}
\end{lstlisting}

\section{Experiment Prompts and Parameters~\label{sec: hyperparameters}}

\subsection{Implementations and Hyperparameters} 
\paragraph{Large Language Models}
Table \ref{tab: detals of models} show the source of LLMs we used in our experiment. We use 1-4 A100s to launch the LLMs with vLLM. The vLLM configuration uses standard configs and use \texttt{fp16} for all models. 

\begin{table}[htbp]
\centering
\resizebox{\textwidth}{!}{\begin{tabular}{ll}
\toprule
Model Name & Model Code/API \\ \midrule
 GPT-3.5-Turbo~\citep{achiam2023gpt}&  Azure api: \texttt{gpt-35-turbo} \\

    Mistral-7b ~\citep{jiang2023mistral} & \texttt{mistralai/Mistral-7B-v0.3} \\

Llama3-8b~\citep{dubey2024llama}& \texttt{meta-llama/Meta-Llama-3-8B-Instruct} \\
Llama3.1-70b~\citep{dubey2024llama}& \texttt{meta-llama/Llama-3.1-70B-Instruct} \\
Deepseek-Coder-6.7b~\citep{guo2024deepseek} & \texttt{deepseek-ai/deepseek-coder-6.7b-instruct} \\
\bottomrule
\end{tabular}}
\caption{Model code/API of our evaluated models.}
\label{tab: detals of models}
\end{table}

\paragraph{Baseline Implementations}

\begin{itemize}
    \item \textbf{PAL}: We use the setting of first generating the entire code and then obtain the answers through a Python environment. We use the implementation from \citet{guo2024deepseek}.
    \item \textbf{Beam Search}: We use the standard implementation provided by vLLM. 
    \item \textbf{Guided Decoding}: We use a widely used implementation from LLM-Reasoners~\citep{hao2023reasoning}. 
    \item \textbf{Self-Consistency}: We reimplement Self-Consistency following~\citep{wang2022self}. For PAL generation, we only weight the answers from successful executed code. 
    \item \textbf{Self Infilling}: We use Self Infilling implementation following ~\citet{zheng2023self}. However this method requires suffix tokens, which is not available for Llama3. Instead, we first generate an entire code and then use logp to heuristically find the start of suffix. 
    \item \textbf{Act}: We use the implementation from AgentBoard \citep{ma2024agentboard}. 
    \item \textbf{ReAct}: We reimplement ReAct following \citet{yao2022react}. Note that we use the setting that LLM could freely choose to alternate between thinking and acting. We find this performs more stably compared to compulsory thinking at each step. 
    \item \textbf{MCTS}: We use the MCTS implementation from LLM-Reasoners~\citep{hao2023reasoning}. However, the original implementation for GSM8K uses subquestions, which is different from other baselines. We unify each step as COT format. 
\end{itemize}

\paragraph{Hyperparameters for Main Experiments}

Table \ref{tab: hyperparameters} shows the hyperparameters used for main experiments for \methodname{}. Most tasks prefer a moderate LLM generation temperature and very low selection temperature. However, we find that Deepseek-Coder prefer otherwise. This is due to Deepseek-Coder generation is more random, making accurate sampling of foresight difficult, therefore we need smaller $\tau$ to prevent overfitting. The FLOPS for \methodname{} is roughly $O(KT_0)$, however we use parallel accelerated implementation and the actual speed is roughtly $O(T_0)$.

For math and coding tasks $N=1$ Sampling baselines (PAL, COT, Self Infilling) use $\alpha=0.6$, $N>1$ use $\alpha=1.0$ if otherwise stated. For agent tasks, Act and ReAct use temperature $\alpha=0.0$. 

Beam Search uses the same beam size and best-of-n as $N$, $N=8$ for math tasks and $N=10$ for coding tasks. The FLOPS for beamsearch is roughly $O(N^2)$ under this implementation.

Guided Decoding uses the standard hyperparameters from LLM-Reasoners implementation: beam size = 5, depth limit = 16, number of generations for each step = 8, beam search temperature 0.5, and reject minimum reward = 0.6. 
 
\begin{table}[!t]
\caption{Hyperparameters for \methodname{} main experiments.}
\centering

\resizebox{\linewidth}{!}{\begin{tabular}{llcc}
\toprule
Method            & Model                                     & Task      & Hyperparameters \\
\midrule
\multirow{12}{*}{\methodname{}} & \multirow{4}{*}{Llama3-8B}                & MATH      &         $\alpha=1.0, \tau=0.01, K=8, T_0=6$         \\
                  &                                           & GSM8K     &   $\alpha=1.0, \tau=0.01, K=8, T_0=6$               \\
                  &                                           & HumanEval &    $\alpha=0.3, \tau=0.05, K=8, T_0=6$             \\
                  &                                           & MBPP      &    $\alpha=1.0, \tau=0.1, K=8, T_0=6$             \\
                  \cmidrule{2-4}
                  & \multirow{2}{*}{Mistral-v0.3}             & MATH      &    $\alpha=1.0, \tau=0.01, K=8, T_0=6$             \\
                  &                                           & GSM8K     &    $\alpha=1.0, \tau=0.01, K=8, T_0=6$             \\
                   \cmidrule{2-4}
                  & \multirow{2}{*}{Deepseek-Coder}           & HumanEval &      $\alpha=0.4, \tau=1.0, K=8, T_0=6$            \\
                  &                                           & MBPP      &    $\alpha=0.4, \tau=1.0, K=8, T_0=6$              \\
                   \cmidrule{2-4}
                  & \multirow{2}{*}{Llama3.1-70B}             & Alfworld  &    $\alpha=1.0, \tau=0.01, K=8, T_0=5$             \\
                  &                                           & PDDL      &     $\alpha=1.0, \tau=0.01, K=8, T_0=5$            \\
                   \cmidrule{2-4}
                  & \multirow{2}{*}{GPT-35-Turbo (azure API)} & Alfworld  &     $\alpha=0.6, \tau=0.01, K=8, T_0=5$            \\
                  &                                           & PDDL      &   $\alpha=0.8, \tau=0.05, K=8, T_0=5$ \\
\bottomrule
\end{tabular}
}
\label{tab: hyperparameters}
\end{table}

\subsection{Prompts}
\input{prompts/code}
\input{prompts/gsm8k_cot}
\input{prompts/gsm8k_pal}
\input{prompts/math_pal}
\input{prompts/alfworld}
\input{prompts/pddl}

%% file: tables/algorithm.tex
\begin{algorithm}
\caption{\methodname{} for Planning}\label{alg:mpc}
\begin{algorithmic}
\State {\bfseries Input:} $\texttt{prompt}$, the language model, maximum number of iterations $T$, sampling number $K$, sampling temperature $\tau$, environment $\texttt{env}$, rollout length $T_0$. 
\State {\bfseries Output:} Action sequence $a_0^{\prime}, \dots a_T^{\prime}$.
\State
\State Set $\mathbf{s}_0 \gets$ Initialize $\texttt{env}$, $\texttt{finish}\gets$ False
\For{$t = 1,2,\dots, T$}
    \State \Comment{Sample Foresight.}
    \State Input \texttt{prompt} and $\mathbf{s}_0^{\prime}, a_0^{\prime},\dots,\mathbf{s}_t^{\prime}$ to language model \texttt{context};
    \For{$k = 1,2,\dots, K$} // \textcolor{gray}{ In parallel}
        \State Sample $a_t^k,\mathbf{s}_{t+1}^k, a_{t+1}^k,\dots, a_{t+T_0}^k, \mathbf{s}_{t+T_0+1}\sim P^\text{LLM}(\cdot \mid \texttt{context})$;
        \State $P_k \gets P^\text{LLM}(\mathbf{a}_{\geq t
        }^k,\mathbf{s}_{> t
        }^k  \mid \texttt{context})$;
        \State $w_k \gets \exp{\left(P_k/\tau\right)}$;
    \EndFor
    \State \Comment{Re-sample based on foresight.}
    \State Sample $j \sim \text { Categorical }\left(\frac{w_1}{\sum_{k=1}^K w_k}, \cdots, \frac{w_K}{\sum_{k=1}^K w_k}\right)$;
    \State Set $a_t^{\prime} \gets a_t^j$;
    \State \Comment{Takes the action $a_t^{\prime}$}
    \State Update $\mathbf{s}_{t+1}, \texttt{finish}\gets$ Execute $a_t^{\prime}$ in environment $\texttt{env}$;
   \State \textbf{break} \textbf{if} $\texttt{finish} \text{ is True}$;
\EndFor
\State \textbf{Return} the action sequence $a_0^{\prime}, \dots a_T^{\prime}$.
\end{algorithmic}
\end{algorithm}

%% file: prompts/code.tex
\begin{tcolorbox}[breakable, title=Prompt Details for Coding Tasks ]
\textbf{\textcolor{blue}{System Prompt}}\\ 
Finish writing the python function. You will only write code blocks. Write 
\# finish after the last line of the function.\\
\begin{tikzpicture}
\draw[dashed] (0,0) -- (\linewidth,0); 
\end{tikzpicture}
\textbf{\textcolor{blue}{Instruction}}\\
\{Instruction of the problem\}
\end{tcolorbox}
\begin{figure}[!ht]
\end{figure}

%% file: prompts/gsm8k_cot.tex
\begin{tcolorbox}[breakable, title=Prompt Details for GSM8K-COT style ]
\textbf{\textcolor{blue}{System Prompt}}\\ 
You will solve math problems following examples.\\
\begin{tikzpicture}
\draw[dashed] (0,0) -- (\linewidth,0); 
\end{tikzpicture}
\textbf{\textcolor{blue}{Examples}}\\
Q: There are 15 trees in the grove. Grove workers will plant trees in the grove today. After they are done, there will be 21 trees. How many trees did the grove workers plant today?

A: There are 15 trees originally. Then there were 21 trees after some more were planted. So there must have been 21 - 15 = 6. The answer is 6.

Q: If there are 3 cars in the parking lot and 2 more cars arrive, how many cars are in the parking lot?

A: There are originally 3 cars. 2 more cars arrive. 3 + 2 = 5. The answer is 5.

...

Q:...

A:...

\begin{tikzpicture}
\draw[dashed] (0,0) -- (\linewidth,0); 
\end{tikzpicture}
\textbf{\textcolor{blue}{Instruction}}\\
Solve this problem following previous examples:
Q: \{question\}
\end{tcolorbox}
\begin{figure}[!h]
\end{figure}

%% file: prompts/gsm8k_pal.tex
\begin{tcolorbox}[breakable, title=Prompt Details for GSM8K-PAL style ]
\textbf{\textcolor{blue}{System Prompt}}\\ 
You will write python program to solve math problems. You will only write code blocks.\\
\begin{tikzpicture}
\draw[dashed] (0,0) -- (\linewidth,0); 
\end{tikzpicture}
\textbf{\textcolor{blue}{Examples}}\\
      Q: Olivia has \$23. She bought five bagels for \$3 each. How much money does she have left?

\# solution in Python:

\begin{lstlisting}[language=Python]
def solution():
    """Olivia has $23. She bought five bagels for $3 each. 
    How much money does she have left?"""
    money_initial = 23
    bagels = 5
    bagel_cost = 3
    money_spent = bagels * bagel_cost
    money_left = money_initial - money_spent
    result = money_left
    return result
\end{lstlisting}

Q: ...

\# solution in Python:

...
\begin{tikzpicture}
\draw[dashed] (0,0) -- (\linewidth,0); 
\end{tikzpicture}
\textbf{\textcolor{blue}{Instruction}}\\
Solve this problem following previous examples:
Q: \{question\}

\# solution in Python:

\begin{lstlisting}[language=Python]
def solution():
\end{lstlisting}
\end{tcolorbox}
\begin{figure}[!h]
\end{figure}

%% file: prompts/math_pal.tex
\begin{tcolorbox}[breakable, title=Prompt Details for MATH-PAL style ]
\textbf{\textcolor{blue}{System Prompt}}\\ 
You will write python program to solve math problems. You will only write code blocks.\\
\begin{tikzpicture}
\draw[dashed] (0,0) -- (\linewidth,0); 
\end{tikzpicture}
\textbf{\textcolor{blue}{Examples}}\\
Let's write python function to solve math problems. You must return the executed result at the end of the function in float. If the final result is an expression, return it in LaTeX in simplest form. You can only write a single function.

Here are some examples:

Question: Find the coefficient of $x^3$ when $3(x^2 - x^3+x) +3(x +2x^3- 3x^2 + 3x^5+x^3) -5(1+x-4x^3 - x^2)$ is simplifie.

solution in Python:

\begin{lstlisting}[language=Python]
from sympy import symbols, simplify

def solution():
    x = symbols('x')
    expr = 3*(x**2 - x**3 + x) + 3*(x + 2*x**3 - 3*x**2 
    \+ 3*x**5 + x**3) - 5*(1 + x - 4*x**3 - x**2)
    simplified_expr = simplify(expr)

    x3_coef = simplified_expr.as_coefficients_dict()[x**3]
    result = x3_coef
    return result
\end{lstlisting}

Q: ...

solution in Python:

...
\begin{tikzpicture}
\draw[dashed] (0,0) -- (\linewidth,0); 
\end{tikzpicture}
\textbf{\textcolor{blue}{Instruction}}\\
Solve this problem following previous examples:
Q: \{question\}

solution in Python:

\begin{lstlisting}[language=Python]
def solution():
\end{lstlisting}
\end{tcolorbox}
\begin{figure}[!h]
\end{figure}

%% file: prompts/alfworld.tex
\begin{tcolorbox}[breakable, title=Prompt Details for AlfWorld ]
\textbf{\textcolor{blue}{System Prompt}}\\ 
You are a helpful assistant. Generate your next step of action after Action. Action must not be empty. e.g. Action: put down cup.\\
\begin{tikzpicture}
\draw[dashed] (0,0) -- (\linewidth,0); 
\end{tikzpicture}
\textbf{\textcolor{blue}{Instruction}}\\
Your task is to interact with a virtual household simulator to accomplish a specific task. With each interaction, you will receive an observation.
Your role is to decide on an action based on the observation. Please ensure that any objects ('{obj}') and receptacles ('{recep}') you mention in your response are present in the observation provided.

Here are the available actions you can take: \\
- take {obj} from {recep} \\
- put {obj} in/on {recep} \\
- open {recep} \\
- close {recep} \\
- toggle {obj}/{recep} \\
- clean {obj} using {recep} \\
- cool {obj} using {recep} \\
- heat {obj} using {recep} \\
- inventory \\
- examine {recep}/{obj} \\
- go to {recep} 

\begin{tikzpicture}
\draw[dashed] (0,0) -- (\linewidth,0); 
\end{tikzpicture}
\textbf{\textcolor{blue}{Examples}}\\
      Your task is to: look at statue under the desklamp. \\
      You are in the middle of a room. Looking quickly around you, you see a coffeetable 1, a diningtable 1, a drawer 4, a drawer 3, a drawer 2, a drawer 1, a dresser 1, a garbagecan 1, a sidetable 2, a sidetable 1, and a sofa 1. \\
      Action: go to dresser 1 \\
      Observation: On the dresser 1, you see a cellphone 3, a newspaper 2, a statue 1, and a television 1. \\
      Action: take statue 1 from dresser 1 \\
      Observation: You pick up the statue 1 from the dresser 1. \\
      Action: go to sidetable 1 \\
      Observation: On the sidetable 1, you see nothing. \\
      Action: go to sidetable 2 \\
      Observation: On the sidetable 2, you see a desklamp 3, a newspaper 1, and a statue 2. \\
      Action: use desklamp 3 \\
      Observation: You turn on the desklamp 3.

\end{tcolorbox}
\begin{figure}[!h]
    \vspace{0.01cm}
\end{figure}

%% file: prompts/pddl.tex
\begin{tcolorbox}[breakable,title=Prompt Details for PDDL ]
\textbf{\textcolor{blue}{System Prompt}}\\ 
You are a master in planning. Generate your next step of action after Action. Action must not be empty. e.g. Action: put down cup.\\
\begin{tikzpicture}
\draw[dashed] (0,0) -- (\linewidth,0); 
\end{tikzpicture}
\textbf{\textcolor{blue}{Instruction}}\\
The robot has four actions: pickup, putdown, stack, and unstack. The domain assumes a world where there are a set of blocks that can be stacked on top of each other, an arm that can hold one block at a time, and a table where blocks can be placed.

The actions defined in this domain include:

pickup  \verb|<block>|: allows the arm to pick up a block from the table if it is clear and the arm is empty. After the pickup action, the arm will be holding the block, and the block will no longer be on the table or clear.

putdown \verb|<block>|: allows the arm to put down a block on the table if it is holding a block. After the putdown action, the arm will be empty, and the block will be on the table and clear.

stack \verb|<block> <block>|: allows the arm to stack a block on top of another block if the arm is holding the top block and the bottom block is clear. After the stack action, the arm will be empty, the top block will be on top of the bottom block, and the bottom block will no longer be clear.

unstack \verb|<block> <block>|: allows the arm to unstack a block from on top of another block if the arm is empty and the top block is clear. After the unstack action, the arm will be holding the top block, the top block will no longer be on top of the bottom block, and the bottom block will be clear.

\begin{tikzpicture}
\draw[dashed] (0,0) -- (\linewidth,0); 
\end{tikzpicture}
\textbf{\textcolor{blue}{Examples}}\\
Goal: The goal is to satisfy the following conditions: b1 is on b2., b2 is on b3.
Observation: b1 is on the table.  b2 is on the table.  B3 is on the table. Robot arm is empty. The b1 is clear. The b2 is clear. The b3 is clear. 

Action: pickup b2

Observation: b1 is on the table.  B2 is on the table.  The b1 is clear. The b3 is clear. You are holding b2.  

Action: stack b2 b3

Observation: b1 is on the table.  b1 is on b2. B3 is on the table. Robot arm is empty. The b1 is clear. The b2 is clear. 

Action: pickup b2. 

Observation: The action is not valid and therefore takes no effect. Please remember to satisfy the restriction of actions. You can also check valid actions. 

Action: check valid actions. 

Observation: valid actions are: pickup b2, unstack b1 b2.

Action: pickup b1

Observation: b2 is on b3. B3 is on the table.  Robot arm is empty. The b2 is clear.  You are holding b1. 

Action: stack b1 b2

Observation: b1 is on b2. b2 is on b3. B3 is on the table.  Robot arm is empty. The b1 is clear. The goal is satisfied.

\end{tcolorbox}
\begin{figure}[!h]
    \vspace{0.01cm}
\end{figure}

%% file: preprint.bbl
\begin{thebibliography}{64}
\providecommand{\natexlab}[1]{#1}
\providecommand{\url}[1]{\texttt{#1}}
\expandafter\ifx\csname urlstyle\endcsname\relax
  \providecommand{\doi}[1]{doi: #1}\else
  \providecommand{\doi}{doi: \begingroup \urlstyle{rm}\Url}\fi

\bibitem[Achiam et~al.(2023)Achiam, Adler, Agarwal, Ahmad, Akkaya, Aleman, Almeida, Altenschmidt, Altman, Anadkat, et~al.]{achiam2023gpt}
Josh Achiam, Steven Adler, Sandhini Agarwal, Lama Ahmad, Ilge Akkaya, Florencia~Leoni Aleman, Diogo Almeida, Janko Altenschmidt, Sam Altman, Shyamal Anadkat, et~al.
\newblock Gpt-4 technical report.
\newblock \emph{arXiv preprint arXiv:2303.08774}, 2023.

\bibitem[Anthony et~al.(2017)Anthony, Tian, and Barber]{anthony2017thinking}
Thomas Anthony, Zheng Tian, and David Barber.
\newblock Thinking fast and slow with deep learning and tree search.
\newblock \emph{Advances in neural information processing systems}, 30, 2017.

\bibitem[Austin et~al.(2021)Austin, Odena, Nye, Bosma, Michalewski, Dohan, Jiang, Cai, Terry, Le, et~al.]{austin2021program}
Jacob Austin, Augustus Odena, Maxwell Nye, Maarten Bosma, Henryk Michalewski, David Dohan, Ellen Jiang, Carrie Cai, Michael Terry, Quoc Le, et~al.
\newblock Program synthesis with large language models.
\newblock \emph{arXiv preprint arXiv:2108.07732}, 2021.

\bibitem[Bachmann \& Nagarajan(2024)Bachmann and Nagarajan]{bachmann2024pitfalls}
Gregor Bachmann and Vaishnavh Nagarajan.
\newblock The pitfalls of next-token prediction.
\newblock \emph{arXiv preprint arXiv:2403.06963}, 2024.

\bibitem[Camacho et~al.(2007)Camacho, Bordons, Camacho, and Bordons]{camacho2007constrained}
Eduardo~F Camacho, Carlos Bordons, Eduardo~F Camacho, and Carlos Bordons.
\newblock \emph{Constrained model predictive control}.
\newblock Springer, 2007.

\bibitem[Chen et~al.(2024{\natexlab{a}})Chen, Monso, Du, Simchowitz, Tedrake, and Sitzmann]{chen2024diffusion}
Boyuan Chen, Diego~Marti Monso, Yilun Du, Max Simchowitz, Russ Tedrake, and Vincent Sitzmann.
\newblock Diffusion forcing: Next-token prediction meets full-sequence diffusion.
\newblock \emph{arXiv preprint arXiv:2407.01392}, 2024{\natexlab{a}}.

\bibitem[Chen et~al.(2021)Chen, Tworek, Jun, Yuan, de~Oliveira~Pinto, Kaplan, Edwards, Burda, Joseph, Brockman, Ray, Puri, Krueger, Petrov, Khlaaf, Sastry, Mishkin, Chan, Gray, Ryder, Pavlov, Power, Kaiser, Bavarian, Winter, Tillet, Such, Cummings, Plappert, Chantzis, Barnes, Herbert-Voss, Guss, Nichol, Paino, Tezak, Tang, Babuschkin, Balaji, Jain, Saunders, Hesse, Carr, Leike, Achiam, Misra, Morikawa, Radford, Knight, Brundage, Murati, Mayer, Welinder, McGrew, Amodei, McCandlish, Sutskever, and Zaremba]{chen2021evaluating}
Mark Chen, Jerry Tworek, Heewoo Jun, Qiming Yuan, Henrique~Ponde de~Oliveira~Pinto, Jared Kaplan, Harri Edwards, Yuri Burda, Nicholas Joseph, Greg Brockman, Alex Ray, Raul Puri, Gretchen Krueger, Michael Petrov, Heidy Khlaaf, Girish Sastry, Pamela Mishkin, Brooke Chan, Scott Gray, Nick Ryder, Mikhail Pavlov, Alethea Power, Lukasz Kaiser, Mohammad Bavarian, Clemens Winter, Philippe Tillet, Felipe~Petroski Such, Dave Cummings, Matthias Plappert, Fotios Chantzis, Elizabeth Barnes, Ariel Herbert-Voss, William~Hebgen Guss, Alex Nichol, Alex Paino, Nikolas Tezak, Jie Tang, Igor Babuschkin, Suchir Balaji, Shantanu Jain, William Saunders, Christopher Hesse, Andrew~N. Carr, Jan Leike, Josh Achiam, Vedant Misra, Evan Morikawa, Alec Radford, Matthew Knight, Miles Brundage, Mira Murati, Katie Mayer, Peter Welinder, Bob McGrew, Dario Amodei, Sam McCandlish, Ilya Sutskever, and Wojciech Zaremba.
\newblock Evaluating large language models trained on code, 2021.

\bibitem[Chen et~al.(2024{\natexlab{b}})Chen, White, Mooney, Payani, Su, and Sun]{chen2024tree}
Ziru Chen, Michael White, Raymond Mooney, Ali Payani, Yu~Su, and Huan Sun.
\newblock When is tree search useful for llm planning? it depends on the discriminator.
\newblock \emph{arXiv preprint arXiv:2402.10890}, 2024{\natexlab{b}}.

\bibitem[Cobbe et~al.(2021)Cobbe, Kosaraju, Bavarian, Chen, Jun, Kaiser, Plappert, Tworek, Hilton, Nakano, Hesse, and Schulman]{cobbe2021gsm8k}
Karl Cobbe, Vineet Kosaraju, Mohammad Bavarian, Mark Chen, Heewoo Jun, Lukasz Kaiser, Matthias Plappert, Jerry Tworek, Jacob Hilton, Reiichiro Nakano, Christopher Hesse, and John Schulman.
\newblock Training verifiers to solve math word problems.
\newblock \emph{arXiv preprint arXiv:2110.14168}, 2021.

\bibitem[Deng et~al.(2020)Deng, Bakhtin, Ott, Szlam, and Ranzato]{deng2020residual}
Yuntian Deng, Anton Bakhtin, Myle Ott, Arthur Szlam, and Marc'Aurelio Ranzato.
\newblock Residual energy-based models for text generation.
\newblock \emph{arXiv preprint arXiv:2004.11714}, 2020.

\bibitem[Du et~al.(2024)Du, Mao, and Tenenbaum]{du2024learning}
Yilun Du, Jiayuan Mao, and Joshua~B Tenenbaum.
\newblock Learning iterative reasoning through energy diffusion.
\newblock \emph{arXiv preprint arXiv:2406.11179}, 2024.

\bibitem[Dubey et~al.(2024)Dubey, Jauhri, Pandey, Kadian, Al-Dahle, Letman, Mathur, Schelten, Yang, Fan, et~al.]{dubey2024llama}
Abhimanyu Dubey, Abhinav Jauhri, Abhinav Pandey, Abhishek Kadian, Ahmad Al-Dahle, Aiesha Letman, Akhil Mathur, Alan Schelten, Amy Yang, Angela Fan, et~al.
\newblock The llama 3 herd of models.
\newblock \emph{arXiv preprint arXiv:2407.21783}, 2024.

\bibitem[Feng et~al.(2024)Feng, Zhang, Gu, Ye, He, and Wang]{feng2024towards}
Guhao Feng, Bohang Zhang, Yuntian Gu, Haotian Ye, Di~He, and Liwei Wang.
\newblock Towards revealing the mystery behind chain of thought: a theoretical perspective.
\newblock \emph{Advances in Neural Information Processing Systems}, 36, 2024.

\bibitem[Freitag \& Al-Onaizan(2017)Freitag and Al-Onaizan]{freitag2017beam}
Markus Freitag and Yaser Al-Onaizan.
\newblock Beam search strategies for neural machine translation.
\newblock \emph{arXiv preprint arXiv:1702.01806}, 2017.

\bibitem[Fu et~al.(2024)Fu, Bailis, Stoica, and Zhang]{fu2024break}
Yichao Fu, Peter Bailis, Ion Stoica, and Hao Zhang.
\newblock Break the sequential dependency of llm inference using lookahead decoding.
\newblock \emph{arXiv preprint arXiv:2402.02057}, 2024.

\bibitem[Gao et~al.(2023)Gao, Madaan, Zhou, Alon, Liu, Yang, Callan, and Neubig]{gao2023pal}
Luyu Gao, Aman Madaan, Shuyan Zhou, Uri Alon, Pengfei Liu, Yiming Yang, Jamie Callan, and Graham Neubig.
\newblock Pal: Program-aided language models.
\newblock In \emph{International Conference on Machine Learning}, pp.\  10764--10799. PMLR, 2023.

\bibitem[Guo et~al.(2024)Guo, Zhu, Yang, Xie, Dong, Zhang, Chen, Bi, Wu, Li, et~al.]{guo2024deepseek}
Daya Guo, Qihao Zhu, Dejian Yang, Zhenda Xie, Kai Dong, Wentao Zhang, Guanting Chen, Xiao Bi, Yu~Wu, YK~Li, et~al.
\newblock Deepseek-coder: When the large language model meets programming--the rise of code intelligence.
\newblock \emph{arXiv preprint arXiv:2401.14196}, 2024.

\bibitem[Ha \& Schmidhuber(2018)Ha and Schmidhuber]{ha2018world}
David Ha and J{\"u}rgen Schmidhuber.
\newblock World models.
\newblock \emph{arXiv preprint arXiv:1803.10122}, 2018.

\bibitem[Hao et~al.(2023)Hao, Gu, Ma, Hong, Wang, Wang, and Hu]{hao2023reasoning}
Shibo Hao, Yi~Gu, Haodi Ma, Joshua~Jiahua Hong, Zhen Wang, Daisy~Zhe Wang, and Zhiting Hu.
\newblock Reasoning with language model is planning with world model.
\newblock \emph{arXiv preprint arXiv:2305.14992}, 2023.

\bibitem[Hendrycks et~al.(2021)Hendrycks, Burns, Kadavath, Arora, Basart, Tang, Song, and Steinhardt]{hendrycks2021measuring}
Dan Hendrycks, Collin Burns, Saurav Kadavath, Akul Arora, Steven Basart, Eric Tang, Dawn Song, and Jacob Steinhardt.
\newblock Measuring mathematical problem solving with the math dataset.
\newblock \emph{arXiv preprint arXiv:2103.03874}, 2021.

\bibitem[Janner et~al.(2022)Janner, Du, Tenenbaum, and Levine]{janner2022planning}
Michael Janner, Yilun Du, Joshua~B Tenenbaum, and Sergey Levine.
\newblock Planning with diffusion for flexible behavior synthesis.
\newblock \emph{arXiv preprint arXiv:2205.09991}, 2022.

\bibitem[Ji et~al.(2023)Ji, Ke, Wang, and Huang]{ji2023language}
Haozhe Ji, Pei Ke, Hongning Wang, and Minlie Huang.
\newblock Language model decoding as direct metrics optimization.
\newblock \emph{arXiv preprint arXiv:2310.01041}, 2023.

\bibitem[Jiang et~al.(2023)Jiang, Sablayrolles, Mensch, Bamford, Chaplot, Casas, Bressand, Lengyel, Lample, Saulnier, et~al.]{jiang2023mistral}
Albert~Q Jiang, Alexandre Sablayrolles, Arthur Mensch, Chris Bamford, Devendra~Singh Chaplot, Diego de~las Casas, Florian Bressand, Gianna Lengyel, Guillaume Lample, Lucile Saulnier, et~al.
\newblock Mistral 7b.
\newblock \emph{arXiv preprint arXiv:2310.06825}, 2023.

\bibitem[Kaplan et~al.(2020)Kaplan, McCandlish, Henighan, Brown, Chess, Child, Gray, Radford, Wu, and Amodei]{kaplan2020scaling}
Jared Kaplan, Sam McCandlish, Tom Henighan, Tom~B Brown, Benjamin Chess, Rewon Child, Scott Gray, Alec Radford, Jeffrey Wu, and Dario Amodei.
\newblock Scaling laws for neural language models.
\newblock \emph{arXiv preprint arXiv:2001.08361}, 2020.

\bibitem[Kurutach et~al.(2018)Kurutach, Clavera, Duan, Tamar, and Abbeel]{kurutach2018model}
Thanard Kurutach, Ignasi Clavera, Yan Duan, Aviv Tamar, and Pieter Abbeel.
\newblock Model-ensemble trust-region policy optimization.
\newblock \emph{arXiv preprint arXiv:1802.10592}, 2018.

\bibitem[Kwon et~al.(2023)Kwon, Li, Zhuang, Sheng, Zheng, Yu, Gonzalez, Zhang, and Stoica]{kwon2023vllm}
Woosuk Kwon, Zhuohan Li, Siyuan Zhuang, Ying Sheng, Lianmin Zheng, Cody Yu, Joey Gonzalez, Hao Zhang, and Ion Stoica.
\newblock vllm: Easy, fast, and cheap llm serving with pagedattention.
\newblock \emph{See https://vllm. ai/(accessed 9 August 2023)}, 2023.

\bibitem[LeCun et~al.(2006)LeCun, Chopra, Hadsell, Ranzato, Huang, et~al.]{lecun2006tutorial}
Yann LeCun, Sumit Chopra, Raia Hadsell, M~Ranzato, Fujie Huang, et~al.
\newblock A tutorial on energy-based learning.
\newblock \emph{Predicting structured data}, 1\penalty0 (0), 2006.

\bibitem[Li et~al.(2024)Li, Liu, Zhou, and Ma]{li2024chain}
Zhiyuan Li, Hong Liu, Denny Zhou, and Tengyu Ma.
\newblock Chain of thought empowers transformers to solve inherently serial problems.
\newblock \emph{arXiv preprint arXiv:2402.12875}, 2024.

\bibitem[Lightman et~al.(2023)Lightman, Kosaraju, Burda, Edwards, Baker, Lee, Leike, Schulman, Sutskever, and Cobbe]{lightman2023let}
Hunter Lightman, Vineet Kosaraju, Yura Burda, Harri Edwards, Bowen Baker, Teddy Lee, Jan Leike, John Schulman, Ilya Sutskever, and Karl Cobbe.
\newblock Let's verify step by step.
\newblock \emph{arXiv preprint arXiv:2305.20050}, 2023.

\bibitem[Lu et~al.(2021)Lu, Welleck, West, Jiang, Kasai, Khashabi, Bras, Qin, Yu, Zellers, et~al.]{lu2021neurologic}
Ximing Lu, Sean Welleck, Peter West, Liwei Jiang, Jungo Kasai, Daniel Khashabi, Ronan~Le Bras, Lianhui Qin, Youngjae Yu, Rowan Zellers, et~al.
\newblock Neurologic a* esque decoding: Constrained text generation with lookahead heuristics.
\newblock \emph{arXiv preprint arXiv:2112.08726}, 2021.

\bibitem[Ma et~al.(2024)Ma, Zhang, Zhu, Yang, Yang, Jin, Lan, Kong, and He]{ma2024agentboard}
Chang Ma, Junlei Zhang, Zhihao Zhu, Cheng Yang, Yujiu Yang, Yaohui Jin, Zhenzhong Lan, Lingpeng Kong, and Junxian He.
\newblock Agentboard: An analytical evaluation board of multi-turn llm agents.
\newblock \emph{arXiv preprint arXiv:2401.13178}, 2024.

\bibitem[Nagabandi et~al.(2018)Nagabandi, Kahn, Fearing, and Levine]{nagabandi2018neural}
Anusha Nagabandi, Gregory Kahn, Ronald~S Fearing, and Sergey Levine.
\newblock Neural network dynamics for model-based deep reinforcement learning with model-free fine-tuning.
\newblock In \emph{2018 IEEE international conference on robotics and automation (ICRA)}, pp.\  7559--7566. IEEE, 2018.

\bibitem[Qin et~al.(2022)Qin, Welleck, Khashabi, and Choi]{qin2022cold}
Lianhui Qin, Sean Welleck, Daniel Khashabi, and Yejin Choi.
\newblock Cold decoding: Energy-based constrained text generation with langevin dynamics.
\newblock \emph{Advances in Neural Information Processing Systems}, 35:\penalty0 9538--9551, 2022.

\bibitem[Qin \& Badgwell(1997)Qin and Badgwell]{qin1997overview}
S~Joe Qin and Thomas~A Badgwell.
\newblock An overview of industrial model predictive control technology.
\newblock In \emph{AIche symposium series}, volume~93, pp.\  232--256. New York, NY: American Institute of Chemical Engineers, 1971-c2002., 1997.

\bibitem[Qin et~al.(2023)Qin, Liang, Ye, Zhu, Yan, Lu, Lin, Cong, Tang, Qian, et~al.]{qin2023toolllm}
Yujia Qin, Shihao Liang, Yining Ye, Kunlun Zhu, Lan Yan, Yaxi Lu, Yankai Lin, Xin Cong, Xiangru Tang, Bill Qian, et~al.
\newblock Toolllm: Facilitating large language models to master 16000+ real-world apis.
\newblock \emph{arXiv preprint arXiv:2307.16789}, 2023.

\bibitem[Racani{\`e}re et~al.(2017)Racani{\`e}re, Weber, Reichert, Buesing, Guez, Jimenez~Rezende, Puigdom{\`e}nech~Badia, Vinyals, Heess, Li, et~al.]{racaniere2017imagination}
S{\'e}bastien Racani{\`e}re, Th{\'e}ophane Weber, David Reichert, Lars Buesing, Arthur Guez, Danilo Jimenez~Rezende, Adri{\`a} Puigdom{\`e}nech~Badia, Oriol Vinyals, Nicolas Heess, Yujia Li, et~al.
\newblock Imagination-augmented agents for deep reinforcement learning.
\newblock \emph{Advances in neural information processing systems}, 30, 2017.

\bibitem[Roziere et~al.(2023)Roziere, Gehring, Gloeckle, Sootla, Gat, Tan, Adi, Liu, Remez, Rapin, et~al.]{roziere2023code}
Baptiste Roziere, Jonas Gehring, Fabian Gloeckle, Sten Sootla, Itai Gat, Xiaoqing~Ellen Tan, Yossi Adi, Jingyu Liu, Tal Remez, J{\'e}r{\'e}my Rapin, et~al.
\newblock Code llama: Open foundation models for code.
\newblock \emph{arXiv preprint arXiv:2308.12950}, 2023.

\bibitem[Shih et~al.(2023)Shih, Sadigh, and Ermon]{shih2023long}
Andy Shih, Dorsa Sadigh, and Stefano Ermon.
\newblock Long horizon temperature scaling.
\newblock In \emph{International Conference on Machine Learning}, pp.\  31422--31434. PMLR, 2023.

\bibitem[Shinn et~al.(2024)Shinn, Cassano, Gopinath, Narasimhan, and Yao]{shinn2024reflexion}
Noah Shinn, Federico Cassano, Ashwin Gopinath, Karthik Narasimhan, and Shunyu Yao.
\newblock Reflexion: Language agents with verbal reinforcement learning.
\newblock \emph{Advances in Neural Information Processing Systems}, 36, 2024.

\bibitem[Shridhar et~al.(2021)Shridhar, Yuan, C\^ot\'e, Bisk, Trischler, and Hausknecht]{ALFWorld20}
Mohit Shridhar, Xingdi Yuan, Marc-Alexandre C\^ot\'e, Yonatan Bisk, Adam Trischler, and Matthew Hausknecht.
\newblock {ALFWorld: Aligning Text and Embodied Environments for Interactive Learning}.
\newblock In \emph{Proceedings of the International Conference on Learning Representations (ICLR)}, 2021.
\newblock URL \url{https://arxiv.org/abs/2010.03768}.

\bibitem[Silver et~al.(2014)Silver, Lever, Heess, Degris, Wierstra, and Riedmiller]{silver2014deterministic}
David Silver, Guy Lever, Nicolas Heess, Thomas Degris, Daan Wierstra, and Martin Riedmiller.
\newblock Deterministic policy gradient algorithms.
\newblock In \emph{International conference on machine learning}, pp.\  387--395. Pmlr, 2014.

\bibitem[Silver et~al.(2017)Silver, Schrittwieser, Simonyan, Antonoglou, Huang, Guez, Hubert, Baker, Lai, Bolton, et~al.]{silver2017mastering}
David Silver, Julian Schrittwieser, Karen Simonyan, Ioannis Antonoglou, Aja Huang, Arthur Guez, Thomas Hubert, Lucas Baker, Matthew Lai, Adrian Bolton, et~al.
\newblock Mastering the game of go without human knowledge.
\newblock \emph{nature}, 550\penalty0 (7676):\penalty0 354--359, 2017.

\bibitem[Smith \& Gelfand(1992)Smith and Gelfand]{smith1992bayesian}
Adrian~FM Smith and Alan~E Gelfand.
\newblock Bayesian statistics without tears: a sampling--resampling perspective.
\newblock \emph{The American Statistician}, 46\penalty0 (2):\penalty0 84--88, 1992.

\bibitem[Snell et~al.(2024)Snell, Lee, Xu, and Kumar]{snell2024scaling}
Charlie Snell, Jaehoon Lee, Kelvin Xu, and Aviral Kumar.
\newblock Scaling llm test-time compute optimally can be more effective than scaling model parameters.
\newblock \emph{arXiv preprint arXiv:2408.03314}, 2024.

\bibitem[Sun et~al.(2024)Sun, Zhuang, Kong, Dai, and Zhang]{sun2024adaplanner}
Haotian Sun, Yuchen Zhuang, Lingkai Kong, Bo~Dai, and Chao Zhang.
\newblock Adaplanner: Adaptive planning from feedback with language models.
\newblock \emph{Advances in Neural Information Processing Systems}, 36, 2024.

\bibitem[Sun \& Yang(2023)Sun and Yang]{sun2023difusco}
Zhiqing Sun and Yiming Yang.
\newblock Difusco: Graph-based diffusion solvers for combinatorial optimization.
\newblock \emph{Advances in Neural Information Processing Systems}, 36:\penalty0 3706--3731, 2023.

\bibitem[Uesato et~al.(2022)Uesato, Kushman, Kumar, Song, Siegel, Wang, Creswell, Irving, and Higgins]{uesato2022solving}
Jonathan Uesato, Nate Kushman, Ramana Kumar, Francis Song, Noah Siegel, Lisa Wang, Antonia Creswell, Geoffrey Irving, and Irina Higgins.
\newblock Solving math word problems with process-and outcome-based feedback.
\newblock \emph{arXiv preprint arXiv:2211.14275}, 2022.

\bibitem[Wang et~al.(2024)Wang, Song, Tian, Peng, Yu, Mi, Su, and Yu]{wang2024litesearch}
Ante Wang, Linfeng Song, Ye~Tian, Baolin Peng, Dian Yu, Haitao Mi, Jinsong Su, and Dong Yu.
\newblock Litesearch: Efficacious tree search for llm.
\newblock \emph{arXiv preprint arXiv:2407.00320}, 2024.

\bibitem[Wang et~al.(2023{\natexlab{a}})Wang, Xie, Jiang, Mandlekar, Xiao, Zhu, Fan, and Anandkumar]{wang2023voyager}
Guanzhi Wang, Yuqi Xie, Yunfan Jiang, Ajay Mandlekar, Chaowei Xiao, Yuke Zhu, Linxi Fan, and Anima Anandkumar.
\newblock Voyager: An open-ended embodied agent with large language models.
\newblock \emph{arXiv preprint arXiv:2305.16291}, 2023{\natexlab{a}}.

\bibitem[Wang et~al.(2023{\natexlab{b}})Wang, Li, Shao, Xu, Dai, Li, Chen, Wu, and Sui]{wang2023math}
Peiyi Wang, Lei Li, Zhihong Shao, RX~Xu, Damai Dai, Yifei Li, Deli Chen, Y~Wu, and Zhifang Sui.
\newblock Math-shepherd: A label-free step-by-step verifier for llms in mathematical reasoning.
\newblock \emph{arXiv preprint arXiv:2312.08935}, 2023{\natexlab{b}}.

\bibitem[Wang \& Zhou(2024)Wang and Zhou]{wang2024chain}
Xuezhi Wang and Denny Zhou.
\newblock Chain-of-thought reasoning without prompting.
\newblock \emph{arXiv preprint arXiv:2402.10200}, 2024.

\bibitem[Wang et~al.(2022)Wang, Wei, Schuurmans, Le, Chi, Narang, Chowdhery, and Zhou]{wang2022self}
Xuezhi Wang, Jason Wei, Dale Schuurmans, Quoc Le, Ed~Chi, Sharan Narang, Aakanksha Chowdhery, and Denny Zhou.
\newblock Self-consistency improves chain of thought reasoning in language models.
\newblock \emph{arXiv preprint arXiv:2203.11171}, 2022.

\bibitem[Wei et~al.(2022)Wei, Wang, Schuurmans, Bosma, Xia, Chi, Le, Zhou, et~al.]{wei2022chain}
Jason Wei, Xuezhi Wang, Dale Schuurmans, Maarten Bosma, Fei Xia, Ed~Chi, Quoc~V Le, Denny Zhou, et~al.
\newblock Chain-of-thought prompting elicits reasoning in large language models.
\newblock \emph{Advances in neural information processing systems}, 35:\penalty0 24824--24837, 2022.

\bibitem[Witkin \& Kass(1988)Witkin and Kass]{witkin1988spacetime}
Andrew Witkin and Michael Kass.
\newblock Spacetime constraints.
\newblock \emph{ACM Siggraph Computer Graphics}, 22\penalty0 (4):\penalty0 159--168, 1988.

\bibitem[Wu et~al.(2023)Wu, Fan, Liu, Zheng, Gong, Jiao, Li, Guo, Duan, Chen, et~al.]{wu2023ar}
Tong Wu, Zhihao Fan, Xiao Liu, Hai-Tao Zheng, Yeyun Gong, Jian Jiao, Juntao Li, Jian Guo, Nan Duan, Weizhu Chen, et~al.
\newblock Ar-diffusion: Auto-regressive diffusion model for text generation.
\newblock \emph{Advances in Neural Information Processing Systems}, 36:\penalty0 39957--39974, 2023.

\bibitem[Wu et~al.(2024{\natexlab{a}})Wu, Morris, and Levine]{wu2024language}
Wilson Wu, John~X Morris, and Lionel Levine.
\newblock Do language models plan ahead for future tokens?
\newblock \emph{arXiv preprint arXiv:2404.00859}, 2024{\natexlab{a}}.

\bibitem[Wu et~al.(2024{\natexlab{b}})Wu, Sun, Li, Welleck, and Yang]{wu2024empirical}
Yangzhen Wu, Zhiqing Sun, Shanda Li, Sean Welleck, and Yiming Yang.
\newblock An empirical analysis of compute-optimal inference for problem-solving with language models.
\newblock \emph{arXiv preprint arXiv:2408.00724}, 2024{\natexlab{b}}.

\bibitem[Xie et~al.(2024)Xie, Kawaguchi, Zhao, Zhao, Kan, He, and Xie]{xie2024self}
Yuxi Xie, Kenji Kawaguchi, Yiran Zhao, James~Xu Zhao, Min-Yen Kan, Junxian He, and Michael Xie.
\newblock Self-evaluation guided beam search for reasoning.
\newblock \emph{Advances in Neural Information Processing Systems}, 36, 2024.

\bibitem[Yao et~al.(2022)Yao, Zhao, Yu, Du, Shafran, Narasimhan, and Cao]{yao2022react}
Shunyu Yao, Jeffrey Zhao, Dian Yu, Nan Du, Izhak Shafran, Karthik Narasimhan, and Yuan Cao.
\newblock React: Synergizing reasoning and acting in language models.
\newblock \emph{arXiv preprint arXiv:2210.03629}, 2022.

\bibitem[Yao et~al.(2024)Yao, Yu, Zhao, Shafran, Griffiths, Cao, and Narasimhan]{yao2024tree}
Shunyu Yao, Dian Yu, Jeffrey Zhao, Izhak Shafran, Tom Griffiths, Yuan Cao, and Karthik Narasimhan.
\newblock Tree of thoughts: Deliberate problem solving with large language models.
\newblock \emph{Advances in Neural Information Processing Systems}, 36, 2024.

\bibitem[Zhao et~al.(2024)Zhao, Ma, Wang, Su, Kong, Xu, Deng, and Yang]{zhao2024empowering}
Haiteng Zhao, Chang Ma, Guoyin Wang, Jing Su, Lingpeng Kong, Jingjing Xu, Zhi-Hong Deng, and Hongxia Yang.
\newblock Empowering large language model agents through action learning.
\newblock \emph{arXiv preprint arXiv:2402.15809}, 2024.

\bibitem[Zheng et~al.(2023)Zheng, Yuan, Zhang, Yang, and Kong]{zheng2023self}
Lin Zheng, Jianbo Yuan, Zhi Zhang, Hongxia Yang, and Lingpeng Kong.
\newblock Self-infilling code generation.
\newblock \emph{arXiv preprint arXiv:2311.17972}, 2023.

\bibitem[Zhou et~al.(2023)Zhou, Yan, Shlapentokh-Rothman, Wang, and Wang]{zhou2023language}
Andy Zhou, Kai Yan, Michal Shlapentokh-Rothman, Haohan Wang, and Yu-Xiong Wang.
\newblock Language agent tree search unifies reasoning acting and planning in language models.
\newblock \emph{arXiv preprint arXiv:2310.04406}, 2023.

\bibitem[Zhuang et~al.(2023)Zhuang, Chen, Yu, Mitra, Bursztyn, Rossi, Sarkhel, and Zhang]{zhuang2023toolchain}
Yuchen Zhuang, Xiang Chen, Tong Yu, Saayan Mitra, Victor Bursztyn, Ryan~A Rossi, Somdeb Sarkhel, and Chao Zhang.
\newblock Toolchain*: Efficient action space navigation in large language models with a* search.
\newblock \emph{arXiv preprint arXiv:2310.13227}, 2023.

\end{thebibliography}
